\documentclass{article}
\usepackage[margin=1in]{geometry}

\usepackage{microtype}
\usepackage{subfigure}
\usepackage{booktabs} % for professional tables
\usepackage[whole]{bxcjkjatype}
\usepackage{hyperref}
\usepackage{amsmath,amsthm,amsfonts,amssymb}
\usepackage{mathtools}
\usepackage{xcolor}
\usepackage{natbib}
\usepackage{algorithm} % for algorithm
\usepackage{algpseudocode} % for algorithm
\usepackage{url}
\usepackage{multirow}
\usepackage{authblk}
\usepackage[T1]{fontenc}
\usepackage[utf8]{inputenc}
\usepackage{bbm}

\def\*#1{\boldsymbol{#1}}

\makeatletter
\def\@endtheorem{\endtrivlist}
\makeatother

\theoremstyle{plain}
\newtheorem*{rep@theorem}{\rep@title}

\newcommand{\newreptheorem}[2]{%
\newenvironment{rep#1}[1]{%
 \def\rep@title{#2 \ref{##1}}%
 \begin{rep@theorem}}%
 {\end{rep@theorem}}}

\newtheorem{theorem}{Theorem}[section]
\newtheorem{lemma}{Lemma}[section]
\newtheorem{corollary}{Corollary}[section]
\newtheorem{assumption}{Assumption}[section]
\newtheorem{definition}{Definition}[section]
\newtheorem{proposition}{Proposition}[section]
\newtheorem{remark}{Remark}[section]

\newreptheorem{theorem}{Theorem}
\newreptheorem{lemma}{Lemma}
\newreptheorem{corollary}{Corollary}
\newreptheorem{assumption}{Assumption}
\newreptheorem{definition}{Definition}
\newreptheorem{proposition}{Proposition}
\newreptheorem{remark}{Remark}

\newcommand{\myPr}{\mathrm{Pr}}

\DeclareMathOperator*{\argmax}{arg\,max}

\newcommand{\lw}[1]{\smash{\lower2.ex\hbox{#1}}}

\newcommand{\RR}{\mathbb{R}}

\newcommand{\EE}{\mathbb{E}}

\newcommand{\cA}{{\cal A}}

\newcommand{\cD}{{\cal D}}

\newcommand{\cF}{{\cal F}}
\newcommand{\cG}{{\cal G}}
\newcommand{\cH}{{\cal H}}

\newcommand{\cN}{{\cal N}}

\newcommand{\cP}{{\cal P}}

\newcommand{\cX}{{\cal X}}

\mathtoolsset{showonlyrefs}
\title{Regret Analysis for \\ Randomized Gaussian Process Upper Confidence Bound}

\author[1,2]{Shion Takeno}
\author[3]{Yu Inatsu}
\author[3]{Masayuki Karasuyama}

\affil[1]{Nagoya University}
\affil[2]{RIKEN AIP}
\affil[3]{Nagoya Institute of Technology}

\affil[ ]{\texttt{takeno.s.mllab.nit@gmail.com}}
\affil[ ]{\texttt{\{inatsu.yu,karasuyama\}@nitech.ac.jp}}
\date{}

\begin{document}
\maketitle

\begin{abstract}%
    Gaussian process upper confidence bound (GP-UCB) is a theoretically established algorithm for Bayesian optimization (BO), where we assume the objective function $f$ follows a GP.
    One notable drawback of GP-UCB is that the theoretical confidence parameter $\beta$ increases along with the iterations and is too large.
    To alleviate this drawback, this paper analyzes the randomized variant of GP-UCB called improved randomized GP-UCB (IRGP-UCB), which uses the confidence parameter generated from the shifted exponential distribution.
    We analyze the expected regret and conditional expected regret, where the expectation and the probability are taken respectively with $f$ and noise and with the randomness of the BO algorithm.
    In both regret analyses, IRGP-UCB achieves a sub-linear regret upper bound without increasing the confidence parameter if the input domain is finite.
    Furthermore, we show that randomization plays a key role in avoiding an increase in confidence parameter by showing that GP-UCB using a constant confidence parameter can incur linearly growing expected cumulative regret.
    Finally, we show numerical experiments using synthetic and benchmark functions and real-world emulators.
\end{abstract}

% Gaussian process upper confidence bound (GP-UCB) is a theoretically established algorithm for Bayesian optimization (BO), where we assume the objective function $f$ follows GP. One notable drawback of GP-UCB is that the theoretical confidence parameter $\beta$ increased along with the iterations is too large. To alleviate this drawback, this paper analyzes the randomized variant of GP-UCB called improved randomized GP-UCB (IRGP-UCB), which uses the confidence parameter generated from the shifted exponential distribution. We analyze the expected regret and conditional expected regret, where the expectation and the probability are taken respectively with $f$ and noises and with the randomness of the BO algorithm. In both regret analyses, IRGP-UCB achieves a sub-linear regret upper bound without increasing the confidence parameter if the input domain is finite. Finally, we show numerical experiments using synthetic and benchmark functions and real-world emulators.

% \begin{keywords}
%   Gaussian process bandits, Bayesian optimization, regret analysis
% \end{keywords}

%%%%%%%%%%%%%%%%%%%%%%%%%%%%%%%%%%%%%%%%%%%%%%%%%%%%%%%%%%%%%%%%%%%%%%%%%%%%%%%%%%%%%%
% \input{manuscripts/HPB-RGPUCB}

\section{Introduction}
\label{sec:intro}

Bayesian optimization (BO) has become a popular black-box optimization method.
BO aims to optimize with fewer function evaluations for an objective function, which is expensive to evaluate.
For this purpose, BO sequentially evaluates the function value at the recommended input, which is chosen by maximizing an acquisition function based on a surrogate Bayesian model.
This paper focuses on BO using a Gaussian process (GP) surrogate model.
GP-based BO has been widely used in many applications, such as automatic machine learning \citep{Snoek2012-Practical}, materials informatics \citep{ueno2016combo}, and drug design \citep{korovina2020-chemBO,griffiths2020-constrained}.

GP upper confidence bound (GP-UCB) \citep{Kushner1964-new,Srinivas2010-Gaussian} is a seminal work for regret-guaranteed BO.
\citet{Srinivas2010-Gaussian} show the sub-linear regret bounds of GP-UCB, which has become a fundamental technique for regret analysis in this field.
On the other hand, GP-UCB suffers from the tuning of the confidence parameter, which is often written as $\beta_t$.
The confidence parameter for the theoretical analysis is too large and needs to increase over time.
Hence, although the analysis of GP-UCB is solid and general, there is still an essential gap between theory and practice.

To resolve this problem, \citet{berk2021-randomized} propose the randomized variants of GP-UCB called RGP-UCB, where the confidence parameter follows the Gamma distribution.
Although \citet{berk2021-randomized} show the superior performance of randomized GP-UCB, their regret analysis appears to contain several technical issues \citep[Appendix~C of][]{takeno2023-randomized}.
We rectified the analysis and showed that GP-UCB with randomized confidence parameters achieves sub-linear expected regret upper bounds under several conditions in the previous conference version of this paper \citep[Section~4.1 and Appendix~D of][]{takeno2023-randomized}.
However, even in this analysis, the confidence parameter is too large and needs to increase over time, as with GP-UCB.

% \red{
% This paper considers the necessity of increasing confidence parameters in GP-UCB-based algorithms.
% %
% As discussed above, increasing confidence parameters is theoretically required but often causes over-exploration, which impairs the practical performance of GP-UCB-based algorithms.
% %
% }

In this paper, we analyze randomized GP-UCB using confidence parameters generated from the shifted exponential distribution, referred to as improved randomized GP-UCB (IRGP-UCB).
First, we show sub-linear expected regret upper bounds of IRGP-UCB, where the regret is averaged over all the randomness of the GP, the noise, and the BO algorithm itself.
In particular, our analysis reveals that IRGP-UCB can achieve a sub-linear expected regret without increasing the confidence parameter when the input domain is finite.
%
% Our analysis tightens the expected regret upper bound, by which increasing the confidence parameter can be avoided if the input domain is finite.
%
On the other hand, since the randomness of the BO algorithm is averaged, the expected regret analysis is not fair in comparing the expected regret upper bounds of deterministic BO algorithms.
That is, depending on the realization of the BO algorithm, the regret averaged by the randomness of GP and the noise may get worse.
Therefore, we further analyze the conditional expected regret averaged over the randomness of the GP and the noise and conditioned on the randomness of the BO algorithm.
Even in this analysis, IRGP-UCB does not require increasing the confidence parameter and maintains the same order of upper bounds.
In addition, although the expectation of the randomized confidence parameters needs to increase, we show that randomized GP-UCB can achieve high-probability regret upper bounds, which implies that randomizing confidence parameters does not diminish the theoretical performance compared with the existing result~\citep{Srinivas2010-Gaussian}.
%
% This analysis can be seen as a generalized version of the analysis of GP-UCB \citep{Srinivas2010-Gaussian}.

The above expected and conditional expected regret analyses raise the question of whether randomization is necessary for achieving a sub-linear expected regret bound.
We show an affirmative answer for this question by showing a problem instance where GP-UCB using a constant confidence parameter incurs linear expected cumulative regret even in finite input domains.
Hence, at least for the expected regret analysis, we can see that randomizing the confidence parameters plays a key role in avoiding increasing confidence parameters.

This paper is organized as follows.
Section~\ref{sec:related_work} reviews the existing regret-guaranteed BO methods in the Bayesian setting, in which the objective function $f$ follows the GP.
Section~\ref{sec:background} provides the technical background.
Section~\ref{sec:algorithm} shows the algorithm of the proposed method.
Section~\ref{sec:theorem} shows the analyses for the expected regret upper bound of IRGP-UCB (Section~\ref{sec:theorem_expected}), the conditional expected regret upper bound of IRGP-UCB (Section~\ref{sec:theorem_conditional_expected}), the high-probability regret upper bound of randomized GP-UCB (Section~\ref{sec:theorem_high_probability}), and the expected regret lower bound of GP-UCB with a constant confidence parameter on a specific problem instance (Section~\ref{sec:theorem_LCB_constant_beta}).
% In Section~\ref{sec:theorem}, we show regret analyses for expected and conditional expected regret.
% %
% In addition, we discuss a high-probability regret bound of the randomized GP-UCB \red{and an expected regret lower bound of the GP-UCB using a constant confidence parameter}.
%
Section~\ref{sec:experiment} shows the wide range of numerical and real-world dataset experiments.
Finally, Section~\ref{sec:conculusion} concludes.

Compared with the previous conference version of this paper \citep{takeno2023-randomized}, we add the following analyses in addition to reorganizing the entire paper.
First, we add the conditional expected regret analysis of IRGP-UCB in Section~\ref{sec:theorem_conditional_expected}.
Second, we show that randomized GP-UCB with increasing confidence parameters can achieve high-probability sub-linear cumulative regret bounds in Section~\ref{sec:theorem_high_probability}.
Third, in Section~\ref{sec:theorem_LCB_constant_beta}, we show that GP-UCB using a constant confidence parameter can incur linear expected cumulative regret.
In addition, we have removed an unnecessary assumption, the kernel's stationarity, from theorems and reorganized Lemma~4.2 (Lemma~4.2 of \citep{takeno2023-randomized}) to improve the lemma's usefulness for the proof.
Finally, we have added one baseline method called the probability of improvement from the maximum of sample path \citep{takeno2024-posterior} in experiments.

\subsection{Related Work}
\label{sec:related_work}

The theoretical assumptions for BO are mainly twofold: the Bayesian setting~\citep{Srinivas2010-Gaussian,Russo2014-learning,Kandasamy2018-Parallelised}, where the objective function $f$ is assumed to be a sample path from a GP, and the frequentist setting~\citep{Srinivas2010-Gaussian,Chowdhury2017-on,janz2020-bandit}, where the objective function $f$ is an element of a reproducing kernel Hilbert space (RKHS).
Hence, we can construct credible intervals directly in the Bayesian setting since $f$ follows the posterior distribution.
On the other hand, although $f$ does not follow the posterior distribution in the frequentist setting, confidence intervals for $f$ based on GP models have been developed~\citep{Srinivas2010-Gaussian,Chowdhury2017-on}.
Based on these tools, we can design BO algorithms that have regret guarantees using GP models.
Note that since a sample path from a GP does not belong to the RKHS with bounded norm, neither analysis subsumes the other (See, e.g., the last paragraph in Section~4 of \citep{Srinivas2010-Gaussian}).
This study considers BO in the Bayesian setting.

Various BO methods have been developed in the literature, for example, expected improvement (EI) \citep{Mockus1978-Application}, entropy search (ES) \citep{Henning2012-Entropy}, and predictive entropy search (PES) \citep{Hernandez2014-Predictive}.
The theoretical analysis of EI for the noiseless case, where function values are not contaminated by noise, has been performed in \citep{vazquez2010convergence,grunewalder2010regret,bull2011convergence} for both Bayesian and frequentist settings.
However, the regret analysis of EI for the noisy case is still limited to the frequentist setting~\citep{wang2014theoretical,tran-the2022regret}.
% Although the regret analysis of EI for the noiseless and frequentist setting, in which $f$ is an element of a reproducing kernel Hilbert space, is provided in \citep{bull2011convergence}, the Bayesian setting has not been considered.
%
Further, although the practical performance of ES and PES has been shown repeatedly, their regret analysis is an open problem.

GP-UCB is one of the prominent studies for theoretically guaranteed BO methods.
\citet{Srinivas2010-Gaussian} show the high-probability bound for cumulative regret.
Furthermore, GP-UCB achieves the sub-linear expected regret upper bounds \citep[Theorem~2.1 and B.1 of][]{takeno2023-randomized}.
However, the confidence parameter must be scheduled as $\beta_t \propto \log t$ in both regret analyses.
Although many studies \citep[for example, ][]{Srinivas2010-Gaussian,Chowdhury2017-on,janz2020-bandit} considered the frequentist setting, this study concentrates on the Bayesian setting.
On the other hand, truncated variance reduction (TRUVAR) \citep{bogunovic2016truncated} achieves high-probability regret bounds.
TRUVAR is based on the variance reduction in potential maximizers determined by UCB and lower confidence bounds.
Hence, TRUVAR may be improved using randomization similar to GP-UCB.

Another well-known regret-guaranteed BO method is Thompson sampling (TS) \citep{Russo2014-learning,Kandasamy2018-Parallelised,takeno2024-posterior}.
TS achieves sub-linear BCR bounds by sequentially observing the optimal point of the GP's posterior sample path.
Although TS does not require any hyperparameter, TS often deteriorates by over-exploration, as discussed in \citep{Shahriari2016-Taking}.
Recently, another posterior sampling-based BO called probability improvement from the maximum of sample path (PIMS) has been proposed.
Known-best expected regret analyses of TS and PIMS are based on the proof technique of our previous conference version of this paper \citep{takeno2023-randomized}.

\citet{Wang2016-Optimization} show regret analysis of the GP estimation (GP-EST) algorithm, which can be interpreted as GP-UCB with the confidence parameter defined using $\hat{m}$, an estimator of $\EE[\max f(\*x)]$.
Their analysis requires an assumption $\hat{m} \geq \EE[\max f(\*x)]$, whose sufficient condition is provided in Corollary~3.5 of \citet{Wang2016-Optimization}.
However, this sufficient condition does not typically hold, as discussed immediately after the corollary in \citet{Wang2016-Optimization}.
Furthermore, the final convergence rate contains $\hat{m}$ itself, whose convergence rate is not clarified.
In contrast, our Lemma~\ref{lem:bound_RGPUCB} shows the bound for $\EE[\max f(\*x)]$ under common regularity conditions.
\citet{Wang2017-Max} show regret analysis of max-value entropy search (MES).
However, it is pointed out that their proof contains several technical issues \citep{takeno2022-sequential}.

\citet{scarlett2018-tight} shows that, under several conditions, there exists an algorithm that achieves a tighter high-probability regret upper bound compared with other well-known algorithms, such as GP-UCB and TS.
However, the algorithm shown by \citet{scarlett2018-tight} depends on the increasing confidence parameters for the construction of confidence bounds.
Therefore, their algorithm also has the same gap between theoretical analysis and practical effectiveness as GP-UCB.

Another relevant line of work is level set estimation (LSE) \citep{gotovos2013-active}.
LSE considers the classification problem of the expensive-to-evaluate black-box functions.
After our preceding conference version of this paper \citep{takeno2023-randomized}, a randomized version of LSE has been proposed by \citet{inatsu2024-active}.
Even for LSE, the randomization avoids increasing the confidence parameter in UCB-based LSE methods.

\section{Background}
\label{sec:background}

% In this section, we describe some background knowledge.

\subsection{Bayesian Optimization}

We consider an optimization problem 
\begin{align*}
    \*x^* = \argmax_{\*x \in \cX} f(\*x),
\end{align*}
where $f$ is a black-box function, $\cX \subset \RR^d$ is an input domain, and $d$ is an input dimension.
Here, black-box means that we can access only function values and cannot obtain other information, such as gradients.
Furthermore, we assume that observations are contaminated by noise.
In addition, we consider the case that $f$ is expensive to evaluate.
Therefore, BO sequentially observes the noisy function value with some policy so that the number of function evaluations is minimized as far as possible.
This policy is determined by an acquisition function (AF) $a: \cX \rightarrow \RR$, which quantifies the utility of function evaluation.
That is, at each iteration $t$, BO queries $\*x_t = \argmax_{\*x \in \cX} a(\*x)$ and obtains $y_t = f(\*x_t) + \epsilon_t$, where $\epsilon_t$ is additive noise.
%
% To determine $\*x_t$, BO is based on acquisition function (AF) as $\*x_t \argmax_{\*x \in \cX} a(\*x)$.
%
Regularity assumptions of $f$ and $\epsilon_t$ are described in Section~\ref{sec:regularity_assumption}.

BO generally employs the GP surrogate model \citep{Rasmussen2005-Gaussian} due to its flexible prediction capability and analytical computations.
The GP surrogate model can be derived under assumptions that $f$ is a sample path from a GP with a zero mean and a kernel function $k: \cX \times \cX \mapsto \RR$, and $\epsilon_t \sim \cN(0, \sigma^2)$ is i.i.d. Gaussian noise with a positive variance $\sigma^2 > 0$.
At each iteration $t$, a dataset $\cD_{t-1} \coloneqq \{ (\*x_i, y_i) \}_{i=1}^{t-1}$ is already obtained from the nature of BO.
Then, the posterior distribution $p(f \mid \cD_{t-1})$ is a GP again.
The posterior mean and variance of $f(\*x)$ are derived as follows:
% $\mu_{t-1}(\*x) = \*k_{t-1}(\*x)^\top \bigl(\*K + \sigma^2 \*I_{t-1} \bigr)^{-1} \*y_{t-1}$ and
% $\sigma_{t-1}^2 (\*x) = k(\*x, \*x) - \*k_{t-1}(\*x) ^\top \bigl(\*K + \sigma^2 \*I_{t-1} \bigr)^{-1} \*k_{t-1}(\*x)$,
\begin{align*}
    \mu_{t-1}(\*x) &= \*k_{t-1}(\*x)^\top \bigl(\*K + \sigma^2 \*I_{t-1} \bigr)^{-1} \*y_{t-1}, \\
    \sigma_{t-1}^2 (\*x) &= k(\*x, \*x) - \*k_{t-1}(\*x) ^\top \bigl(\*K + \sigma^2 \*I_{t-1} \bigr)^{-1} \*k_{t-1}(\*x),
\end{align*}
where $\*k_{t-1}(\*x) \coloneqq \bigl( k(\*x, \*x_1), \dots, k(\*x, \*x_{t-1}) \bigr)^\top \in \RR^{t-1}$, $\*K \in \RR^{(t-1)\times (t-1)}$ is the kernel matrix whose $(i, j)$-element is $k(\*x_i, \*x_j)$, $\*I_{t-1} \in \RR^{(t-1)\times (t-1)}$ is the identity matrix, and $\*y_{t-1} \coloneqq (y_1, \dots, y_{t-1})^\top \in \RR^{t-1}$.
Hereafter, we denote that the probability density function (PDF) $p(\cdot \mid \cD_{t-1}) = p_t (\cdot)$, the probability $\Pr (\cdot \mid \cD_{t-1}) = \myPr_t (\cdot) $, and the expectation $\EE[\cdot \mid \cD_{t-1}] = \EE_t[\cdot]$ for brevity.

\subsection{Regularity Assumptions}
\label{sec:regularity_assumption}

We assume that the GP surrogate model is correct.
That is, $\epsilon_t \sim \cN(0, \sigma^2)$ is i.i.d. Gaussian noise with a positive variance $\sigma^2 > 0$, and the underlying true function $f$ is a sample path from a known GP with a zero mean and a kernel function $k$.
Here, the zero mean is without loss of generality \citep{Rasmussen2005-Gaussian}.
In addition, we assume that the kernel function $k$ is normalized.
Therefore, the kernel $k(\*x, \*x^\prime)$ satisfies the condition $k(\*x, \*x^\prime) \leq 1$ for all $\*x, \*x^\prime \in \cX$.

Furthermore, when $\cX$ is continuous, we consider the following smoothness assumption:
\begin{assumption}
    Let $\cX \subset [0, r]^d$ be a compact and convex set, where $r > 0$.
    Assume that the kernel $k$ satisfies the following condition on the derivatives of a sample path $f$.
    There exists the constants $a, b > 0$ such that,
    \begin{align*}
        \Pr \left( \sup_{\*x \in \cX} \left| \frac{\partial f}{\partial \*x_j} \right| > L \right) \leq a \exp \left( - \left(\frac{L}{b}\right)^2 \right),\text{ for } j \in [d],
    \end{align*}
    where $[d] = \{1, \dots, d\}$.
    \label{assump:continuous_X}
\end{assumption}
This assumption holds for stationary \citep{genton2001-classes} and four times differentiable kernels \citep[Theorem~5 of]{Ghosal2006-posterior}, such as RBF kernel and Mat\'{e}rn kernels with a parameter $\nu ][] 2$, as discussed in Section~4 of \citet{Srinivas2010-Gaussian}.
Therefore, this assumption is commonly used, for example, in \citet{Srinivas2010-Gaussian,Kandasamy2018-Parallelised,paria2020-flexible,takeno2024-posterior}.

\subsection{Regret}

In this paper, we evaluate the performance of BO methods by \emph{cumulative regret} \citep{Srinivas2010-Gaussian,Russo2014-learning,Kandasamy2018-Parallelised,paria2020-flexible,takeno2024-posterior}.
We incur the instantaneous regret $r_t = f(\*x^*) - f(\*x_t)$ for each function evaluation.
The cumulative regret for $T$ iterations is defined as the sum of instantaneous regrets over $T$ function evaluations, defined as follows:
\begin{align*}
    {\rm R}_T = \sum_{t=1}^T r_t.
\end{align*}
In particular, BO methods that achieve sub-linear cumulative regret (i.e., such that $\lim_{T \rightarrow \infty} R_T / T = 0$) are called \emph{no-regret}.
No-regret property also indicates that the best regret $\min_{t \in [T]} r_t$ converges to $0$ since $\min_{t \in [T]} r_t < R_T / T$.
Therefore, the main goal of theoretical analysis is to show sub-linearity of cumulative regret.
Note that regret is a random quantity due to $f$, $\{\epsilon_t\}_{t \geq 1}$, and the randomness of the BO algorithm.

\paragraph{Maximum information gain}: 
The convergence rates of regret bounds are characterized by \emph{maximum information gain} (MIG) \citep{Srinivas2010-Gaussian,vakili2021-information}.
MIG $\gamma_T$ is defined as follows:
\begin{definition}[Maximum information gain]
    Let $f \sim \cG\cP (0, k)$.
    Let $A = \{ \*a_i \}_{i=1}^T \subset \cX$.
    Let $\*f_A = \bigl(f(\*a_i) \bigr)_{i=1}^T$, $\*\epsilon_A = \bigl(\epsilon_i \bigr)_{i=1}^T$, where $\forall i, \epsilon_i \sim \cN(0, \sigma^2)$, and $\*y_A = \*f_A + \*\epsilon_A \in \RR^T$.
    Then, MIG $\gamma_T$ is defined as follows:
    \begin{align*}
        \gamma_T \coloneqq \max_{A \subset \cX; |A| = T} I(\*y_A ; \*f_A),
    \end{align*}
    where $I$ is the Shannon mutual information.
\end{definition}
MIG is known to be sub-linear for commonly used kernel functions, for example, $\gamma_T = O\bigl( (\log T)^{d+1} \bigr)$ for RBF kernels and $\gamma_T = O\bigl( T^{\frac{d}{2\nu + d}} (\log T)^{\frac{2\nu}{2\nu + d}} \bigr)$ for Mat\`{e}rn-$\nu$ kernels \citep{Srinivas2010-Gaussian,vakili2021-information}.

\subsection{GP-UCB}
\label{sec:GPUCB}

GP-UCB selects the next input by maximizing UCB as follows:
\begin{align*}
    \*x_t = \argmax_{\*x \in \cX} \mu_{t-1}(\*x) + \beta^{1/2}_t \sigma_{t-1}(\*x).
\end{align*}
In the existing regret analyses~\citep{Srinivas2010-Gaussian,takeno2023-randomized}, to obtain the sub-linear cumulative regret bounds, the confidence parameter $\beta_t$ must be scheduled as $\beta_t = \Theta \bigl( \log(|\cX| t) \bigr)$ for finite input domains.

\section{Algorithm}
\label{sec:algorithm}

Algorithm \ref{alg:RGPUCB} shows the pseudo-code of IRGP-UCB.
% IRGP-UCB is a simple procedure shown in Algorithm \ref{alg:RGPUCB}.
%
The difference from the original GP-UCB is that $\{\zeta_t\}$ is a sequence of random variables, not a monotonically increasing (non-random) sequence.
Using the generated $\zeta_t$, the next input is chosen as follows:
\begin{align*}
    \*x_t = \argmax_{\*x \in \cX} \mu_{t-1}(\*x) + \zeta^{1/2}_t \sigma_{t-1}(\*x).
\end{align*}
Our choice of the distribution of $\zeta_t$ is a \emph{shifted exponential distribution} \citep[for example, ][]{beg1980estimation,lam1994estimation}, whose PDF is written as follows:
\begin{align*}
    p(\zeta ; s_t, \lambda) = 
    \begin{cases}
    \lambda \exp\bigl( - \lambda (\zeta - s_t) \bigr) & \text{if } \zeta \geq s_t, \\
    0 & \text{otherwise},
    \end{cases}
\end{align*}
where $\lambda$ is a rate parameter.
This can be seen as the distribution of the sum of $s_t$ and $Z \sim {\rm Exp} (\lambda)$.
Thus, sampling from the shifted exponential distribution can be easily performed using an exponential distribution as follows:
\begin{align}
    \zeta_t \gets s_t + Z_t, \text{ where } Z_t \sim {\rm Exp} (\lambda).
    \label{eq:sampling_beta}
\end{align}
The theoretical choice of $\{ s_t \}_{t \geq 1}$ and $\lambda$ will be shown in Section~\ref{sec:theorem}.

\begin{algorithm}[!t]
    \caption{IRGP-UCB}\label{alg:RGPUCB}
    \begin{algorithmic}[1]
        \Require Input space $\cX$, Parameters $\{ s_t \}_{t \geq 1}$ and $\lambda$ for $\zeta_t$, GP prior $\mu=0$ and $k$
        \State $\cD_{0} \gets \emptyset $
        \For{$t = 1, \dots$}
            \State Fit GP to $\cD_{t-1}$
            \State Generate $\zeta_t$ by Eq.~\eqref{eq:sampling_beta}
            \State $\*x_t \gets \argmax_{\*x \in \cX} \mu_{t-1}(\*x) + \zeta_t^{1/2} \sigma_{t-1}(\*x)$
            \State Observe $y_t = f(\*x_t) + \epsilon_t$ and $\cD_{t} \gets \cD_{t-1} \cup (\*x_t, y_t)$
        \EndFor
    \end{algorithmic}
\end{algorithm}

\section{Regret Analysis}
\label{sec:theorem}

In this section, we analyze the performance of IRGP-UCB and the necessity of randomizing confidence parameters.
First, we analyze the average-case performance through expected regret averaged over all the randomness of $f$, $\{\epsilon_t\}_{t \geq 1}$, and $\{\zeta_t\}_{t \geq 1}$.
%
% Regarding expected regret, IRGP-UCB achieves sub-linear upper bounds without increasing the confidence parameter when the input domain is finite.
%
Second, we analyze conditional expected regret averaged over $f$ and $\{\epsilon_t\}_{t \geq 1}$ and conditioned on $\{\zeta_t\}_{t \geq 1}$.
%
% This analysis shows that IRGP-UCB does not require increasing the confidence parameter when the input domain is finite, even if we consider \red{high-probability bound} regarding the randomness of the BO algorithm.
%
Third, we derive high-probability regret bounds of randomized GP-UCB with increasing confidence parameters.
Lastly, we show that GP-UCB using a constant confidence parameter can incur linearly growing expected cumulative regret.

%%%%%%%%%%%%%%%%%%%%%%%%%%%%%%%%%%%%%%%%%%%%%%%%%%%%%%%%%%%%%%%%%%%%%%%%%%%%%%%%%%%%%%%%%%%%%%
\subsection{Expected Regret Bounds}
\label{sec:theorem_expected}
%%%%%%%%%%%%%%%%%%%%%%%%%%%%%%%%%%%%%%%%%%%%%%%%%%%%%%%%%%%%%%%%%%%%%%%%%%%%%%%%%%%%%%%%%%%%%%
Here, we derive the expected cumulative regret bound, which is also called Bayesian cumulative regret (BCR) \citep{Russo2014-learning}.
The definition of BCR is
\begin{align}
    {\rm BCR}_T \coloneqq \EE\left[\sum_{t=1}^T r_t \right], \label{eq:BCR}
\end{align}
where the expectation is taken with all the randomness, that is, $f$, $\{\epsilon_t\}_{t \geq 1}$, and $\{\zeta_t\}_{t \geq 1}$.

First, for completeness, we provide a modified version of the basic lemma derived in \citep{Srinivas2010-Gaussian}:
\begin{lemma}
    Suppose that $f$ is a sample path from a GP with zero mean and a positive semidefinite kernel $k$, and $\cX$ is finite.
    Pick $\delta \in (0, 1)$ and $t \geq 1$.
    Then, for any given $\cD_{t-1}$,
    \begin{align*}
        \myPr_t \left( f(\*x) \leq \mu_{t-1}(\*x) + \beta^{1/2}_{\delta} \sigma_{t-1}(\*x), \forall \*x \in \cX \right)
        \geq 1 - \delta,
    \end{align*}
    where $\beta_{\delta} = 2 \log (|\cX| /( 2 \delta))$.
    \label{lem:bound_srinivas}
\end{lemma}
See Appendix~\ref{app:proof_srinivas} for the proof.
This lemma differs slightly from Lemma~5.1 of \citet{Srinivas2010-Gaussian}, since, in Lemma~\ref{lem:bound_srinivas}, the iteration $t$ is fixed, and $\beta_{\delta}$ does not depend on $t$.

Based on Lemma~\ref{lem:bound_srinivas}, we show the following lemma, which will play a key role in the subsequent analyses:
\begin{lemma}
    Let $f \sim \cG \cP (0, k)$, where $k$ is a positive semidefinite kernel, and $\cX$ be finite.
    Assume that $\zeta_t$ follows a shifted exponential distribution with $s_t = 2 \log (|\cX| / 2)$ and $\lambda = 1/2$.
    Then, for any given $\cD_{t-1}$, the following inequality holds:
    \begin{align*}
        \EE_t[f(\*x^*)] \leq \EE_t \left[\max_{\*x \in \cX} \mu_{t-1}(\*x) + \zeta^{1/2}_t \sigma_{t-1}(\*x) \right],
    \end{align*}
    for all $t \geq 1$.
    \label{lem:bound_RGPUCB}
\end{lemma}
\begin{proof}
    We here show a short proof of Lemma~\ref{lem:bound_RGPUCB}, although a detailed proof is shown in Appendix~\ref{app:proof_lemma_IRGPUCB}.
    %
    % From the tower property of the expectation, it suffices to show the following inequality:
    % \begin{align*}
    %     \EE_t[f(\*x^*)] \leq \EE_t \left[\max_{\*x \in \cX} \mu_{t-1}(\*x) + \zeta^{1/2}_t \sigma_{t-1}(\*x) \right].
    % \end{align*}
    % %
    % Since this inequality only considers the expectation given $\cD_{t-1}$, we can fix $\cD_{t-1}$ in the proof.
    %
    From Lemma~\ref{lem:bound_srinivas}, we can obtain the following inequality:
    \begin{align*}
        % &1 - \delta \leq F_t\left( \max_{\*x \in \cX} \mu_{t-1}(\*x) + \beta^{1/2}_{\delta} \sigma_{t-1}(\*x) \right), \\
        % &\Leftrightarrow 
        F_t^{-1}(1 - \delta) \leq \max_{\*x \in \cX} \mu_{t-1}(\*x) + \beta^{1/2}_{\delta} \sigma_{t-1}(\*x),
    \end{align*}
    where $F_t (\cdot) \coloneqq \myPr_t\left( f(\*x_*) < \cdot \right)$ is a cumulative distribution function of $f(\*x_*)$, and $F_t^{-1}$ is its inverse function.
    Then, substituting $U \sim {\rm Uni}(0, 1)$ into $\delta$ and taking the expectation, we obtain the following inequality:
    \begin{align*}
        \EE_t \left[ f(\*x_*) \right] \leq \EE_U \left[ \max_{\*x \in \cX} \mu_{t-1}(\*x) + \beta^{1/2}_{U} \sigma_{t-1}(\*x) \right],
    \end{align*}
    which can be derived in a similar way to inverse transform sampling.
    Hence, $\beta^{1/2}_{U} = 2 \log (|\cX| / 2) - 2 \log(U)$ results in a random variable, which follows the shifted exponential distribution.
    Consequently, we can obtain the desired bound.
\end{proof}

% It is worth noting that the above proof technique is versatile and requires only the UCB of $f(\*x_*)$.
% %
% The above proof is mainly based on the general fact $\EE_X [X] \leq \EE_U [{\rm UCB}_X (1 - U)]$, where $X$ is an arbitrary random variable and ${\rm UCB}_X$ is its UCB.
% %
% Therefore, roughly speaking, if we obtain the bound that corresponds Lemma~\ref{lem:bound_srinivas} for an objective function in other practical settings of GP-UCB extensions, such as multi-objective BO \citep{paria2020-flexible}, we can perform a similar derivation to Lemma~\ref{lem:bound_RGPUCB}.
%
% Therefore, if we know the UCB of $f(\*x_*)$, we can construct the randomized UCB, whose expectation becomes the upper bound of $\EE[f(\*x_*)]$ even for other practical settings of GP-UCB extensions, such as multi-objective BO \citep{paria2020-flexible}.
%
% Therefore, this technique can apply to more complex problem settings, in which GP-UCB-based methods have been extended, if we know the UCB of the maximum value.

Lemma~\ref{lem:bound_RGPUCB} shows a direct upper bound of the expectation of $f(\*x^*)$.
% based on randomized UCB $v_t(\*x) \coloneqq \mu_{t-1}(\*x) + \zeta^{1/2}_t \sigma_{t-1}(\*x)$.
A notable consequence of Lemma~\ref{lem:bound_RGPUCB} is an upper bound of BCR as follows:
\begin{align*}
    {\rm BCR}_T 
    &= \sum_{t=1}^T \EE\left[ f(\*x^*) - v_t(\*x_t) \right] + \EE\left[ v_t(\*x_t) - f(\*x_t) \right] \\
    &= \sum_{t=1}^T \EE \left[ \EE_t \left[ f(\*x^*) - v_t(\*x_t) \right] \right] + \EE\left[ v_t(\*x_t) - f(\*x_t) \right] \\
    &\leq \sum_{t=1}^T \EE\left[ v_t(\*x_t) - f(\*x_t) \right],
\end{align*}
where $v_t(\*x) \coloneqq \mu_{t-1}(\*x) + \zeta^{1/2}_t \sigma_{t-1}(\*x)$.
The above inequality is obtained by the selection rule of IRGP-UCB.
%
% This upper bound eliminates the first term, which cannot be bounded without increasing the confidence parameter in the conventional regret decomposition Eq.~\eqref{eq:regret_depomsition}. 
%
% Therefore, we can bound BCR as 
% \begin{align*}
%     {\rm BCR}_T 
%     &= \sum_{t=1}^T \EE\left[ f(\*x^*) - v_t(\*x_t) \right] + \EE\left[ v_t(\*x_t) - f(\*x_t) \right] \\
%     &\leq \sum_{t=1}^T \EE\left[ v_t(\*x_t) - f(\*x_t) \right],
% \end{align*}
% which is a notable difference from the regret decomposition with $u_t(\*x)$.
%
Importantly, $\EE[\zeta_t] = 2 + s_t$ is a constant since $s_t = 2\log(|\cX| / 2)$ does not depend on $t$ in Lemma~\ref{lem:bound_RGPUCB}.
Thus, using Lemma~\ref{lem:bound_RGPUCB}, we can obtain the BCR bound for finite input domains:
\begin{theorem}[BCR bound for finite domain]
    Let $f \sim \cG \cP (0, k)$, where $k$ is a positive semidefinite kernel and $k(\*x, \*x) \leq 1$, and $\cX$ be finite.
    Assume that $\zeta_t$ follows a shifted exponential distribution with $s_t = 2 \log (|\cX| / 2)$ and $\lambda = 1/2$ for any $t \geq 1$.
    Then, by running IRGP-UCB with $\zeta_t$, BCR can be bounded as follows: 
    \begin{align*}
        {\rm BCR}_T \leq \sqrt{C_1 C_2 T \gamma_T},
    \end{align*}
    where $C_1 \coloneqq 2 / \log(1 + \sigma^{-2})$ and $C_2 \coloneqq 2 + 2 \log (|\cX| / 2)$. 
    \label{theo:BCR_IRGPUCB_discrete}
\end{theorem}
See Appendix~\ref{app:proof_discrete} for detailed proof.
% %
% \paragraph{Proof Sketch:}
%     By using the randomized UCB $V_t (\*x) \coloneqq \mu_{t-1}(\*x) + \zeta^{1/2}_t \sigma_{t-1}(\*x)$, since $\*x_t = \argmax_{\*x \in \cX} V_t(\*x)$, we can obtain
%     \begin{align*}
%         {\rm BCR}_T 
%         &= \EE \left[ \sum_{t=1}^T  f(\*x^*) - V_t(\*x_t) + V_t(\*x_t) - f(\*x_t) \right] \\
%         &\leq \EE \left[ \sum_{t=1}^T V_t(\*x_t) - f(\*x_t) \right] \quad \bigl(\because \text{Lemma~\ref{lem:bound_RGPUCB}} \bigr) \\
%         &= \EE \left[ \sum_{t=1}^T \zeta^{1/2}_t \sigma_{t-1}(\*x_t) \right] \\
%         &\leq \sqrt{ \sum_{t=1}^T \EE \left[ \zeta_t \right] C_1 \gamma_T} \\
%         &= \sqrt{ C_1 C_2 T \gamma_T},
%     \end{align*}
%     where we use several basic inequalities in the second inequality.
%     %
%     See Appendix~\ref{app:proof_discrete} for detailed proof.

Theorem~\ref{theo:BCR_IRGPUCB_discrete} has an important implication that IRGP-UCB does not need to schedule the parameters $s_t$ and $\lambda$ of $\zeta_t$, in contrast to GP-UCB and RGP-UCB.
%
% First, Theorem~\ref{theo:BCR_IRGPUCB_discrete} shows the convergence rate $O(\sqrt{T \gamma_T})$, in which the multiplicative $\sqrt{\log T}$ factor is improved compared with $O(\sqrt{T \gamma_T \log T})$ achieved by GP-UCB \citep{Srinivas2010-Gaussian} and RGP-UCB.
%
% Thus, IRGP-UCB achieves the best convergence rate for BCR.
%
% Second, more importantly, IRGP-UCB does not need to schedule the parameters $s_t$ and $\lambda$ of $\zeta_t$ in contrast to GP-UCB and RGP-UCB.
%
Therefore, through randomization, IRGP-UCB essentially alleviates the problem that the well-known $\beta_t \propto \log(t)$ strategy results in a practically too large confidence parameter.
Further, over-exploration in the later iterations can be avoided.
IRGP-UCB is the first GP-UCB-based method that enjoys the above practical benefits without sacrificing regret guarantee.

On the other hand, note that the $\sqrt{\log |\cX|}$ dependence remains as with the prior works \citep{Srinivas2010-Gaussian,Russo2014-learning,Kandasamy2018-Parallelised,paria2020-flexible}.
In BO, we are usually interested in the case that the total number of iterations $T < |\cX|$.
Therefore, in such case, although our regret upper bound $O\bigl(\sqrt{\gamma_T T \log |\cX|}\bigr)$ is slightly better than the existing one $O\bigl(\sqrt{\gamma_T T \log (|\cX|T)}\bigr)$, the difference between those regret upper bounds is only the constant term.

% It is worth noting that our key lemma (Lemma~\ref{lem:bound_RGPUCB}) mainly requires Lemma~\ref{lem:bound_srinivas} only, which is expected to be satisfied in a wide range of exploration problems in which a GP is used as a surrogate model. 
% %
% Therefore, we conjecture that the same proof technique can be applicable to more advanced problem settings, such as multi-objective BO \citep{paria2020-flexible}, for which further analyses are important future directions of our results.

Next, we show the BCR bound for the infinite (continuous) domain:
\begin{theorem}[BCR bound for infinite domain]
    Let $f \sim \cG \cP (0, k)$, where $k$ is a positive semidefinite kernel, $k(\*x, \*x) \leq 1$, and Assumption~\ref{assump:continuous_X} holds.
    Assume that $\zeta_t$ follows a shifted exponential distribution with $s_t = 2d \log(bdr t^2 \bigl( \sqrt{\log (ad)} + \sqrt{\pi} / 2\bigr)) - 2 \log 2$ and $\lambda = 1/2$ for any $t \geq 1$.
    Then, by running IRGP-UCB, BCR can be bounded as follows: 
    \begin{align*}
        {\rm BCR}_T \leq \frac{\pi^2}{6} + \sqrt{C_1 T \gamma_T (2 + s_T)},
    \end{align*}
    where $C_1 \coloneqq 2 / \log(1 + \sigma^{-2})$. 
    \label{theo:BCR_IRGPUCB_continuous}
\end{theorem}
See Appendix~\ref{app:proof_continuous} for the proof.
Unfortunately, in this case, $\EE[\zeta_t] = O(\log t)$ and the resulting BCR bound is $O(\sqrt{T \gamma_T \log T})$.
This is because the discretization of input domain $\cX_t$, updated as $|\cX_t| = O(t^{2d})$, is required to obtain the BCR bound for the infinite domain.
Since we cannot avoid the dependence on $|\cX|$ also in Theorem~\ref{theo:BCR_IRGPUCB_discrete}, the resulting BCR bound in Theorem~\ref{theo:BCR_IRGPUCB_continuous} requires the term $\log(|\cX_t|) = O\bigl(2d \log(t) \bigr)$.

\subsection{Conditional Expected Regret Bounds}
\label{sec:theorem_conditional_expected}
%%%%%%%%%%%%%%%%%%%%%%%%%%%%%%%%%%%%%%%%%%%%%%%%%%%%%%%%%%%%%%%%%%%%%%%%%%%%%%%%%%%%
In this section, we analyze the conditional expected regret of IRGP-UCB.
This analysis is newly provided compared with the previous conference version of our paper~\citep{takeno2023-randomized}.
The regret depends on the randomness of $f$, $\{\epsilon_t\}_{t \geq 1}$, and $\{\zeta_t\}_{t \geq 1}$.
Although the randomness of $f$ and $\{\epsilon_t\}_{t \geq 1}$ arise from the problem setup, $\{\zeta_t\}_{t \geq 1}$ is produced by the user.
Hence, distinguishing the randomness, we analyze the following conditional expected regret:
\begin{align*}
    \EE_{f, \{\epsilon_t\}_{t \geq 1}}[ f(\*x^*) - f(\*x_t) \mid \{\zeta_t\}_{t \geq 1}],
\end{align*}
where the expectation is taken with $f$ and $\{\epsilon_t\}_{t \geq 1}$.
Note that $\EE_{f, \{\epsilon_t\}_{t \geq 1}}[ f(\*x^*) - f(\*x_t) \mid \{\zeta_t\}_{t \geq 1}]$ is a random variable depending on $\{\zeta_t\}_{t \geq 1}$.
Hence, showing the sub-linear high-probability upper bound of the conditional expected regret implies that the BO algorithm achieves a good performance, averaged over a random environment, with a high-probability concerning the randomness of the algorithm.

Let $\cP$ be a set of random variables included in the problem setup ($\cP = \{f, \{ \epsilon_t \}_{t \geq 1}\}$) and $\cA$ be a set of random variables included in the algorithm ($\cA = \{ \zeta_t \}_{t \geq 1}$ in the case of IRGP-UCB).
In the usual expected regret analysis of the deterministic BO algorithm, such as GP-UCB \citep[for example,][]{takeno2023-randomized,paria2020-flexible}, the expectation over the problem setup $\EE_{\cP} [R_T]$ is considered.
When we interpret the confidence parameter in GP-UCB as a random variable equal to $\beta_t$ with probability 1 ($\cA = \{ \beta_t \}_{t \geq 1}$), 
$\EE_{\cP} [R_T]$
is equivalent to the joint expectation
$\EE_{\cP, \cA} [R_T]$.
Further, we obviously see that the conditional expected regret is also equivalent: $\EE_{\cP, \cA} [R_T] = \EE_{\cP} [R_T \mid \cA]$.
On the other hand, since IRGP-UCB is stochastic, possible definitions of the expected regret are
$\EE_{\cP, \cA} [R_T]$ 
and 
$\EE_{\cP} [R_T \mid \cA]$, but these two regrets can be different.
Theorems~\ref{theo:BCR_IRGPUCB_discrete} and \ref{theo:BCR_IRGPUCB_continuous} show the theoretical rate of 
$\EE_{\cP, \cA} [R_T]$.
However, these theorems do not guarantee anything about the conditional expected regret
$\EE_{\cP} [R_T \mid \cA]$.
The following remark provides a motivation for deriving a conditional expected regret bound:
\begin{remark}
    Although 
    $\EE_{\cP, \cA} [R_T]$ 
    derived by Theorems~\ref{theo:BCR_IRGPUCB_discrete} and \ref{theo:BCR_IRGPUCB_continuous} reveal average behavior of the regret of IRGP-UCB, it obscures variability caused by $\{ \zeta_t \}_{t \geq 1}$.
    Hence, comparing 
    $\EE_{\cP} [R_T]$ 
    of GP-UCB and 
    $\EE_{\cP, \cA} [R_T]$ 
    of IRGP-UCB can be unfair.
    On the other hand, $\EE_{\cP} [R_T \mid \cA]$ is a random variable, whose randomness stems from $\{\zeta_t\}_{t \geq 1}$.
    Therefore, by showing a high-probability bound of 
    $\EE_{\cP} [R_T \mid \cA]$, 
    we can 1) fairly compare the regret of IRGP-UCB with $\EE_{\cP} [R_T]$ of GP-UCB in a sense that the expectation is taken over the same random variables $\cP$, and 2) demonstrate that the expected regret of IRGP-UCB can be stably bounded with respect to the random behavior of the algorithm. 
\end{remark}

First, we analyze the case of finite $\cX$.
We show the following theorem:
\begin{theorem}
    Let $f \sim \cG \cP (0, k)$, where $k$ is a positive semidefinite kernel and $k(\*x, \*x) \leq 1$, and $\cX$ be finite.
    Assume that $\zeta_t$ follows a shifted exponential distribution with $s_t = 2 \log (|\cX| / 2)$ and $\lambda = 1/2$ for any $t \geq 1$.
    Then, by running IRGP-UCB with $\zeta_t$, the conditional expected regret can be bounded with probability at least $1 - \delta$ as follows: 
    \begin{align*}
        % &\Pr \left( \EE_{f, \{\epsilon_t\}_{t \geq 1}} \left[ \sum_{t=1}^T f(\*x^*) - f(\*x_t) \mid \{\zeta_t\}_{t \geq 1} \right] \leq U(T, \delta) \right) \geq 1 - \delta, \forall T \geq 1, \\
        %
        &\Pr \left( \forall T \geq 1, \EE_{f, \{\epsilon_t\}_{t \geq 1}} \left[ \sum_{t=1}^T f(\*x^*) - f(\*x_t) \mid \{\zeta_t\}_{t \geq 1} \right] \leq U \bigl(T, \delta \bigr) \right) \geq 1 - \delta, 
    \end{align*}
    where 
    \begin{align*}
        U(T, \delta) 
        &= 6\sqrt{T \log\left(\frac{\pi^2 T^2}{3\delta} \right)} + \sqrt{C_1 \gamma_T \left( T s_T + T + 2\sqrt{T \log\left(\frac{\pi^2 T^2}{3\delta} \right)} + 2\log\left(\frac{\pi^2 T^2}{3\delta} \right)\right) } \\
        % &= O \left( \sqrt{T \gamma_T \log|\cX| + \gamma_T \sqrt{T \log(1/\delta)}} + \gamma_T \log(1 / \delta) \right),
    \end{align*}
    and $C_1 \coloneqq 2 / \log(1 + \sigma^{-2})$.
    \label{theo:HPCER_IRGPUCB_discrete}
\end{theorem}
\begin{proof}
    We show the proof sketch here only.
    See Appendix~\ref{app:proof_conditional_hpcr_discrete} for the details.

    The first key point of the proof is transforming the conditional expected regret using Lemma~\ref{lem:bound_RGPUCB} and the properties of the expectation as follows:
    % We can obtain the upper bound by the properties of the expectation and Lemma~\ref{lem:bound_RGPUCB}:
    \begin{align*}
        & \sum_{t=1}^T \EE_{f, \{\epsilon_t\}_{t \geq 1}} \left[  f(\*x^*) - f(\*x_t) \mid \{\zeta_t\}_{t \geq 1} \right] \\
        &= \sum_{t=1}^T \EE_{ \cD_{t-1}} \left[  \EE_{f \mid \cD_{t-1}} \left[  f(\*x^*) \right] -  \mu_{t-1}(\*x_t) \mid \{\zeta_{i} \}_{i \leq t} \right] \\
        % &= \sum_{t=1}^T \EE_{ \cD_{t-1}} \left[ 
        %         \EE_{f \mid \cD_{t-1}} \left[  f(\*x^*) \right] 
        %         - \EE_{\zeta_t \mid \cD_{t-1}} \left[ v_t(\*x_t) \right] + \EE_{\zeta_t \mid \cD_{t-1}} \left[ v_t(\*x_t) \right]
        %         - v_t(\*x_t) + v_t(\*x_t)
        %         -  \mu_{t-1}(\*x_t) 
        %     \mid \{\zeta_{i} \}_{i \leq t} \right] \\
        &\leq \sum_{t=1}^T \EE_{ \cD_{t-1}} \left[ 
                \EE_{\zeta_t \mid \cD_{t-1}} \left[ v_t(\*x_t) \right]
                - v_t(\*x_t) + \zeta_t^{1/2} \sigma_{t-1}(\*x_t)
            \mid \{\zeta_{i} \}_{i \leq t} \right] \qquad (\because \text{Lemma~\ref{lem:bound_RGPUCB}}) \\
        &=
            \underbrace{\sum_{t=1}^T  \EE_{ \cD_{t-1}, \zeta_t} \left[ v_t(\*x_t) \mid \{\zeta_{i} \}_{i < t} \right]
            - \EE_{ \cD_{t-1}} \left[ v_t(\*x_t) \mid \{\zeta_{i} \}_{i \leq t} \right]}_{A_1}
            + \underbrace{\EE \left[ \sum_{t=1}^T \zeta_t^{1/2} \sigma_{t-1}(\*x_t) \mid \{\zeta_{t} \}_{t > 1} \right]}_{A_2},
    \end{align*}
    where $v_t(\*x_t) = \mu_{t-1}(\*x_t) + \zeta_t^{1/2} \sigma_{t-1}(\*x_t)$.
    Then, $A_2$ can be bounded from above using MIG and the high-probability upper bound of chi-square random variables \citep{Laurent2000-adaptive} shown in Lemma~\ref{lem:Laurent}.

    The second technical key point is bounding $A_1$ by Azuma's inequality~\citep{azuma1967weighted}, which requires the conditions that $A_1$ is a sum of the martingale difference sequence with conditional subgaussian property (See Lemma~\ref{lem:azuma} for details).
    We show that $A_1$ satisfies this condition using several well-known lemmas shown in Appendix~\ref{app:proof_conditional_hpcr_discrete}.
    %
    % Furthermore, we can show that $A_1$ is a sum of the martingale sequence adapted to the history $\cH_{t} = \{ \zeta_i \}_{i \leq t}$.
    % %
    % Therefore, we can apply Azuma's inequality \citep{azuma1967weighted}.
    %
    Finally, we can obtain the desired result by applying the union bound for the upper bounds of $A_1$ and $A_2$ for all $T \geq 1$.
\end{proof}

Theorem~\ref{theo:HPCER_IRGPUCB_discrete} shows the same benefits as Theorem~\ref{theo:BCR_IRGPUCB_discrete}.
First, under the above mild condition $\min \{s_T^2 / 2, s_T\} T > \log\bigl(\pi^2 T^2 /  (3\delta) \bigr)$, the dominant term in $U(T, \delta)$ is $\sqrt{2 C_1 T \gamma_T s_T}$, which is the same as Theorem~\ref{theo:BCR_IRGPUCB_discrete}.
Therefore, Theorem~\ref{theo:HPCER_IRGPUCB_discrete} maintains the rate $O\bigl(\sqrt{T \gamma_T \log (|\cX|}) \bigr)$ under mild condition.
Second, the distribution of $\zeta_t$ is the same as in Theorem~\ref{theo:BCR_IRGPUCB_discrete}.
Hence, increasing the confidence parameter is also avoided even if we consider the high-probability bound regarding the randomness of the algorithm.

Next, as with the expected regret bounds, we extend Theorem~\ref{theo:HPCER_IRGPUCB_discrete} to the continuous case:
\begin{theorem}
    Let $f \sim \cG \cP (0, k)$, where $k$ is a positive semidefinite kernel and $k(\*x, \*x) \leq 1$, and Assumption~\ref{assump:continuous_X} holds.
    Assume that $\zeta_t$ follows a shifted exponential distribution with $s_t = 2d \log(bdr t^2 \bigl( \sqrt{\log (ad)} + \sqrt{\pi} / 2\bigr)) - 2 \log 2$ and $\lambda = 1/2$ for any $t \geq 1$.
    Then, by running IRGP-UCB with $\zeta_t$, the conditional expected regret can be bounded with probability at least $1 - \delta$ as follows: 
    \begin{align*}
        &\Pr \left( \forall T \geq 1, \EE_{f, \{\epsilon_t\}_{t \geq 1}} \left[ \sum_{t=1}^T f(\*x^*) - f(\*x_t) \mid \{\zeta_t\}_{t \geq 1} \right] \leq U \bigl(T, \delta \bigr) \right) \geq 1 - \delta, 
    \end{align*}
    where
    \begin{align*}
        U(T, \delta) 
        &= \frac{\pi^2}{6} + 6\sqrt{T \log\left(\frac{\pi^2 T^2}{3\delta} \right)} 
        % \\ &\quad 
        + \sqrt{C_1 \gamma_T \left( T s_T + T + 2\sqrt{T \log\left(\frac{\pi^2 T^2}{3\delta} \right)} + 2\log\left(\frac{\pi^2 T^2}{3\delta} \right)\right) },
    \end{align*}
    and $C_1 \coloneqq 2 / \log(1 + \sigma^{-2})$. 
    \label{theo:HPCER_IRGPUCB_continuous}
\end{theorem}
See Appendix~\ref{app:proof_conditional_hpcr_continuous} for the proof.
The resulting upper bound maintains the same rate as the expected regret bound shown in Theorem~\ref{theo:BCR_IRGPUCB_continuous} if $\min \{s_T^2 / 2, s_T\} T > 2 \log\bigl(\pi^2 T^2 /  (3\delta) \bigr)$.
\subsection{High-Probability Regret Bounds}
\label{sec:theorem_high_probability}

In this section, we consider the high-probability regret bounds of the randomized GP-UCB, which is newly added compared with the previous conference version of this paper.
For simplicity, we focus on the case of the finite $\cX$.
As discussed in \citep{Russo2014-learning}, a direct consequence of Markov's inequality is
\begin{align*}
    \Pr \left(R_T \leq {\rm BCR}_T / \delta \right) \geq 1 - \delta,
\end{align*}
for any $\delta \in (0, 1)$.
Therefore, the following proposition can be easily obtained:
\begin{proposition}
    Assume the same condition as in Theorem~\ref{theo:BCR_IRGPUCB_discrete}.
    Then, by running IRGP-UCB, for some fixed $T \geq 1$, the following holds:
    \begin{align*}
        \Pr \left(R_T \leq \sqrt{C_1 C_2 T \gamma_T} / \delta \right) \geq 1 - \delta,
    \end{align*}
    where $C_1 \coloneqq 2 / \log(1 + \sigma^{-2})$ and $C_2 \coloneqq 2 + 2 \log (|\cX| / 2)$.
    \label{prop:markov_conclusion}
\end{proposition}

However, this proposition shows the regret bound for some fixed $T \geq 1$.
Ideally, an anytime algorithm that guarantees that the sub-linear regret bound holds for all $T \geq 1$ with high-probability is desired.
For example, the analysis of GP-UCB \citep{Srinivas2010-Gaussian} is anytime and shows regret upper bound $O(\sqrt{T \gamma_T \log(T |\cX| /\delta)})$.
In addition, the resulting rate $O(1/\delta)$ regarding $\delta$ is worse than that of GP-UCB $O(\log(1/\delta))$.
Therefore, we focus on obtaining an anytime regret upper bound with the $\log(1/\delta)$ dependence regarding $\delta$.
%
%
% However, we can show that by increasing $\zeta_t$ similarly to the original GP-UCB, IRGP-UCB achieves the high-probability regret bounds.

First, we consider high-probability bounds using the randomized confidence parameter.
Since the randomized confidence parameter $\zeta_t$ is scheduled as $\EE[\zeta_t] = \Theta(\log t)$, we can derive the following lemma as with Lemma 5.1 in \citep{Srinivas2010-Gaussian}:
%
% As with Lemma 5.1 in \citep{Srinivas2010-Gaussian}, we can show the following lemma:
\begin{lemma}
    Let $f \sim \cG \cP (0, k)$, where $k$ is a positive semidefinite kernel, and $\cX$ be finite.
    Pick $\delta \in (0, 1)$.
    Assume that $\zeta_t = 2 \log ( 1 / U_t )$, where $\{ U_t \}_{t \geq 1}$ is a sequence of mutually independent random variables that satisfy $\EE[U_t] = \frac{6 \delta}{|\cX| t^2 \pi^2}$ and $\Pr(U_t \in (0, 1)) = 1$ for all $t \geq 1$.
    Then, the following inequality holds:
    \begin{align*}
        \Pr \left( \forall t \geq 1, \forall \*x \in \cX, |f(\*x) - \mu_{t-1}(\*x)| \leq \zeta^{1/2}_t \sigma_{t-1}(\*x) \right) 
        \geq 1 - \delta.
    \end{align*}
    \label{lem:high_prob_bound_RGPUCB}
\end{lemma}
\begin{proof}
    % We here show the short proof of Lemma~\ref{lem:high_prob_bound_RGPUCB} although detailed proof is shown in Appendix~\ref{app:proof_high_prob_bound_RGPUCB}.
    %
    As with Lemma 5.1 in \citep{Srinivas2010-Gaussian}, for any $t \geq 1$, $\*x \in \cX$, $\cD_{t-1}$, and $U_t$, we obtain 
    \begin{align*}
        \myPr \left( |f(\*x) - \mu_{t-1}(\*x)| > \zeta^{1/2}_t \sigma_{t-1}(\*x) \mid U_t, \cD_{t-1} \right) \leq U_t.
    \end{align*}
    Taking the expectation with respect to $\{ U_t \}_{t \geq 1}$ and $\cD_{t-1}$, we obtain
    \begin{align*}
        \Pr \left( |f(\*x) - \mu_{t-1}(\*x)| > \zeta^{1/2}_t \sigma_{t-1}(\*x) \right) 
        \leq \EE[U_t].
    \end{align*}
    Then, by applying the union bound for all $t \geq 1$ and $\*x \in \cX$, we obtain
    \begin{align*}
        \Pr \left( \forall t \geq 1, \forall \*x \in \cX, |f(\*x) - \mu_{t-1}(\*x)| \leq \zeta^{1/2}_t \sigma_{t-1}(\*x) \right) 
        \geq 1 - |\cX| \sum_{t=1}^T \EE[U_t] 
        \geq 1 - \delta.
    \end{align*}
\end{proof}
This analysis can be seen as a generalized version of the analysis of GP-UCB since the randomized confidence parameter is equivalent to GP-UCB if we set $U_t = \frac{6 \delta}{|\cX| t^2 \pi^2}$ almost surely.

Then, by combining Lemma~\ref{lem:high_prob_bound_RGPUCB} and existing analysis techniques, we can obtain the following theorems:
\begin{theorem}
    Let $f \sim \cG \cP (0, k)$, where $k$ is a positive semidefinite kernel and $k(\*x, \*x) \leq 1$, and $\cX$ be finite.
    Pick $\delta \in (0, 1)$.
    %
    % Assume that $\zeta_t$ follows a shifted exponential distribution with $s_t = 2 \log (|\cX| t^2 \pi^2 / (12\delta))$ and $\lambda = 1/2$ for any $t \geq 1$.
    Assume that $\zeta_t = 2 \log ( 1 / U_t )$, where $\{ U_t \}_{t \geq 1}$ is a sequence of mutually independent random variables that satisfy $\EE[U_t] = \frac{6 \delta}{|\cX| t^2 \pi^2}$ and $\Pr(U_t \in (0, 1)) = 1$ for all $t \geq 1$.
    Then, by running the randomized GP-UCB with $\zeta_t$, the cumulative regret can be bounded with probability at least $ 1 - \delta$ as follows: 
    \begin{align*}
        &\Pr \left( \forall T \geq 1, R_T \leq 2 \sqrt{C_1 \gamma_T \sum_{t=1}^T \zeta_t } \right) \geq 1 - \delta, 
    \end{align*}
    where $C_1 \coloneqq 2 / \log(1 + \sigma^{-2})$.
    % \begin{align*}
    %     R_T &\leq 
    %     2 \sqrt{ C_1 \gamma_T \left( 2T s_T + 2T + 2\sqrt{2T \log\left( \frac{T^2 \pi^2}{3\delta}\right)} + 2\log\left( \frac{T^2 \pi^2}{3 \delta}\right)  \right) }\\
    %     &= O \bigl( \sqrt{T \gamma_T \log (|\cX| T / \delta)} \bigr).
    % \end{align*}
    \label{theo:HPCR_general_GPUCB_discrete}
\end{theorem}
See Appendix~\ref{app:proof_hpcr} for the proof.
Note that the resulting upper bound contains the random quantity $\zeta_t$.
Thus, this theorem still does not rigorously imply a sub-linear regret upper bound.
On the other hand, we can see that Theorem~\ref{theo:HPCR_general_GPUCB_discrete} is a generalized version of the analysis of GP-UCB \citep{Srinivas2010-Gaussian} since the randomized GP-UCB is equivalent to GP-UCB if $U_t = \frac{6 \delta}{|\cX| t^2 \pi^2}$ almost surely.
If we adopt Theorem~\ref{theo:HPCR_general_GPUCB_discrete} to IRGP-UCB and additionally take the upper bound regarding $\sum_{t=1}^T \zeta_t$, we can obtain the following theorem:
\begin{corollary}
    Let $f \sim \cG \cP (0, k)$, where $k$ is a positive semidefinite kernel and $k(\*x, \*x) \leq 1$, and $\cX$ be finite.
    Pick $\delta \in (0, 1)$.
    Assume that $\zeta_t$ follows a shifted exponential distribution with $s_t = 2 \log (|\cX| t^2 \pi^2 / (6 \delta))$ and $\lambda = 1/2$ for any $t \geq 1$.
    Then, by running IRGP-UCB with $\zeta_t$, the cumulative regret can be bounded with probability at least $ 1 - \delta$ as follows: 
    \begin{align*}
        &\Pr \left( \forall T \geq 1, R_T \leq 
        2 \sqrt{C_1 \gamma_T \left( T s_T + T + 2\sqrt{T \log\left( \frac{\pi^2 T^2}{3\delta}\right)} + 2\log\left( \frac{\pi^2 T^2}{3 \delta}\right)  \right) } \right) \geq 1 - \delta, 
    \end{align*}
    where $C_1 \coloneqq 2 / \log(1 + \sigma^{-2})$.
    \label{theo:HPCR_IRGPUCB_discrete}
\end{corollary}
See Appendix~\ref{app:proof_hpcr} for the proof.
We can see that even if confidence parameters are randomized, the same order regret upper bound as that of GP-UCB can be obtained.
Thus, this result implies that randomization does not deteriorate the high-probability regret upper bounds compared with the existing result shown by \citet{Srinivas2010-Gaussian}.
%
% To our knowledge, this is the first high-probability cumulative regret upper bound of the randomized GP-UCB-based algorithm.

However, in Theorem~\ref{theo:HPCR_general_GPUCB_discrete}, we can see that
\begin{align*}
    \EE[\zeta_t] 
    = \EE[ - 2 \log (U_t) ]
    \geq -2 \log ( \EE[U_t]) 
    = 2 \log\left( \frac{|\cX| t^2 \pi^2}{3 \delta} \right).
\end{align*}
Therefore, although $\zeta_t$ may be smaller than the usual $\beta_t$ in GP-UCB, its expectation $\EE[\zeta_t]$ is larger than the usual $\beta_t$. 
We leave the high-probability regret analysis without increasing the confidence parameter as future work.

\subsection{Expected Regret of GP-UCB with Constant Confidence Parameter}
\label{sec:theorem_LCB_constant_beta}

In this section, we show an expected regret lower bound of GP-UCB with any constant confidence parameter.
This analysis is added from the previous conference version of this paper.
We can show that GP-UCB, using any constant confidence parameter, incurs linear expected cumulative regret in the following simple problem instance:
\begin{theorem}
    Assume that $\cX = \{ \*x^{(1)}, \*x^{(2)} \}$, $\epsilon_t \sim \cN(0, 1)$ for all $t \geq 1$, and 
    \begin{align*}
        \left(
            \begin{array}{c}
                 f(\*x^{(1)}) \\
                 f(\*x^{(2)})
            \end{array}
        \right) \sim \cN \left(
            \left(
            \begin{array}{c}
                 0 \\
                 0
            \end{array}
        \right),
        \left(
            \begin{array}{cc}
                 1 & \rho \\
                 \rho & 0.99
            \end{array}
        \right)
        \right),
    \end{align*}
    where $\rho < 1$ is a covariance parameter.
    Then, if GP-UCB with any constant confidence parameter $\beta_t^{(1/2)} = c > 0$ runs, then BCR grows linearly, that is,
    \begin{align*}
        {\rm BCR}_T &= \EE \left[ \sum_{t=1}^T f(\*x^*) - f(\*x_t) \right] = \Omega(T).
    \end{align*}
    \label{theo:GPUCB_BCR_LB}
\end{theorem}
See Appendix~\ref{app:proof_GPUCB_BCR_LB} for the proof.
In the proof, we show that the probability of the extreme case, where the GP-UCB algorithm continue to evaluate $\*x^{(1)}$ even if $f(\*x^{(1)}) < f(\*x^{(2)})$, is rigorously positive.

Theorem~\ref{theo:GPUCB_BCR_LB} suggests that, under assumptions of Theorems~\ref{theo:BCR_IRGPUCB_discrete} and \ref{theo:HPCER_IRGPUCB_discrete}, the confidence parameter must be increased in the GP-UCB algorithm to obtain sub-linear expected regret upper bounds.
Therefore, randomizing confidence parameters as in IRGP-UCB is necessary to avoid an increase in confidence parameters.

\section{Experiments}
\label{sec:experiment}

We demonstrate the experimental results on synthetic and benchmark functions and the materials dataset provided in \citep{liang2021benchmarking}.
As a baseline, we performed EI \citep{Mockus1978-Application}, TS \citep{Russo2014-learning}, MES \citep{Wang2017-Max}, joint entropy search (JES) \citep{hvarfner2022-joint}, GP-UCB \citep{Srinivas2010-Gaussian}, and PIMS \citep{takeno2024-posterior}.
For the posterior sampling in TS, PIMS, MES, and JES, we used random Fourier features \citep{Rahimi2008-Random}.
For Monte Carlo estimation in MES and JES, we used ten samples.
We evaluate the practical performance of BO by simple regret $f(\*x^*) - \max_{t \leq T} f(\*x_t)$.

\begin{figure}[t]
    \centering
    \includegraphics[width=0.45\linewidth]{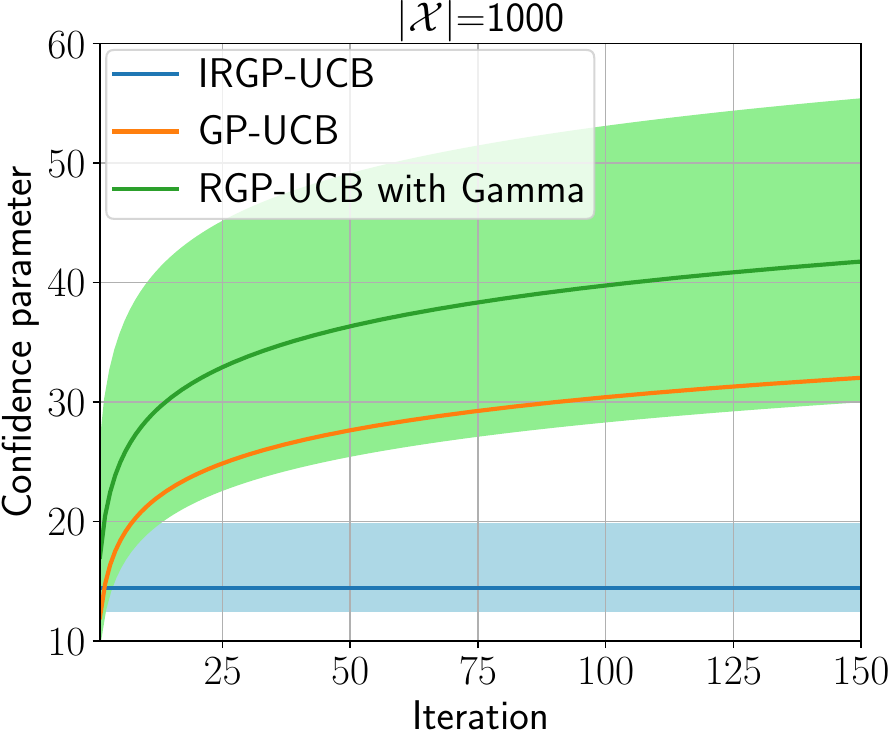}
    \caption{
        This figure shows the confidence parameters of GP-UCB-based methods.
        For GP-UCB, the solid line represents $\beta_t$.
        For RGP-UCB and IRGP-UCB, the solid lines and shaded areas represent the expectations $\EE[\zeta_t]$ and $95 \%$ credible intervals, respectively.
        %
        % The right shows the average and standard errors of simple regrets.
        }
    \label{fig:synthetic-confidence-param}
\end{figure}

\begin{figure}[t]
    \centering
    \includegraphics[width=0.45\linewidth]{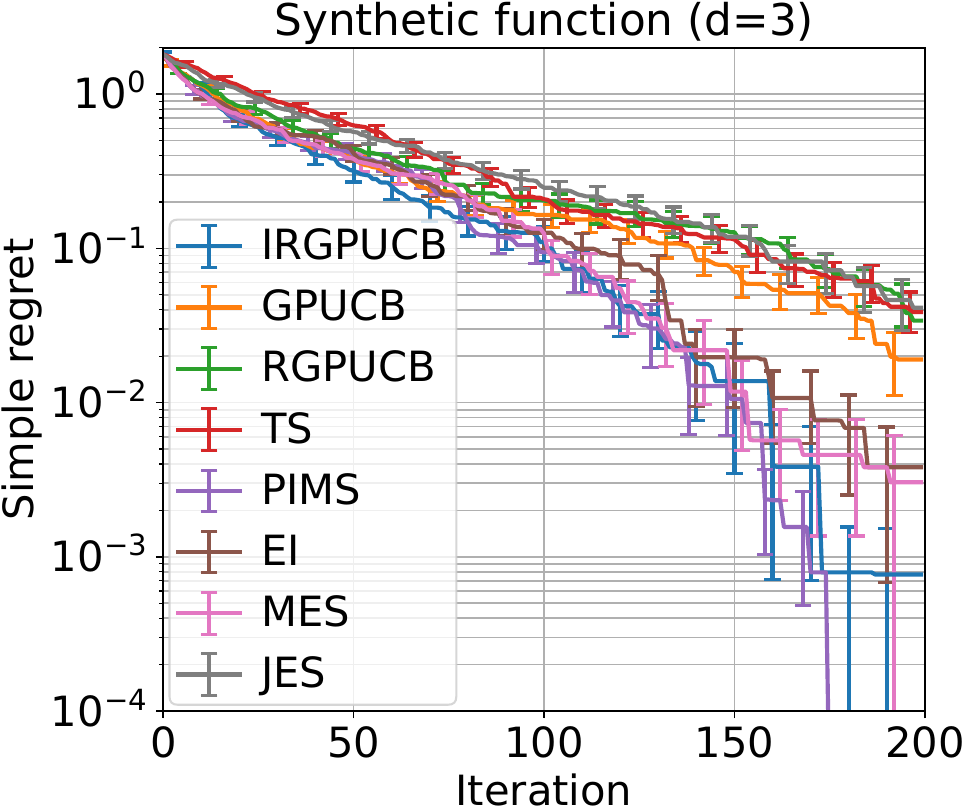}
    \caption{
        This figure shows the average and standard errors of simple regrets.
        }
    \label{fig:synthetic-regret}
\end{figure}

\subsection{Synthetic Function Experiment}

We use synthetic functions generated from $\cG \cP (0, k)$, where $k$ is a Gaussian kernel with a length scale parameter $\ell = 0.1$ and the input dimension $d = 3$.
We set the noise variance $\sigma^2 = 10^{-4}$.
The input domain consists of equally divided points in $[0, 0.9]$, that is, $\cX = \{0, 0.1, \dots, 0.9\}^d$ and $|\cX| = 1000$.
Figure~\ref{fig:synthetic-confidence-param} shows the theoretical confidence parameters of GP-UCB, RGP-UCB with ${\rm Gamma}(\kappa_t=\log(|\cX|t^2) / \log(1.5), \theta = 1)$, and IRGP-UCB with a two-parameter exponential distribution.
We can see that the confidence parameters for GP-UCB and RGP-UCB are considerably large due to a logarithmic increase, particularly in the later iterations.
In contrast, the confidence parameter of IRGP-UCB is drastically smaller than that of GP-UCB and does not change with respect to the iteration.
Therefore, we can observe that IRGP-UCB alleviates the over-exploration.

Next, we report the simple regret in Figure~\ref{fig:synthetic-regret}.
Ten synthetic functions and ten initial training datasets are randomly generated.
Thus, the average and standard error for $10 \times 10$ trials are reported.
The hyperparameters for GP are fixed to those used to generate the synthetic functions.
The confidence parameters for GP-UCB, RGP-UCB, and IRGP-UCB are set as in Figure~\ref{fig:synthetic-confidence-param}.
We can see that the regrets of GP-UCB, RGP-UCB, TS, and JES converge slowly.
We conjecture that this slow convergence came from over-exploration.
On the other hand, EI, PIMS, MES, and IRGP-UCB show fast convergence.
In particular, IRGP-UCB achieves the best average in most iterations.
These results suggest that IRGP-UCB bridges the gap between theory and practice in contrast to GP-UCB and RGP-UCB.

\begin{figure}[!t]
    \centering
    \includegraphics[width=0.325\linewidth]{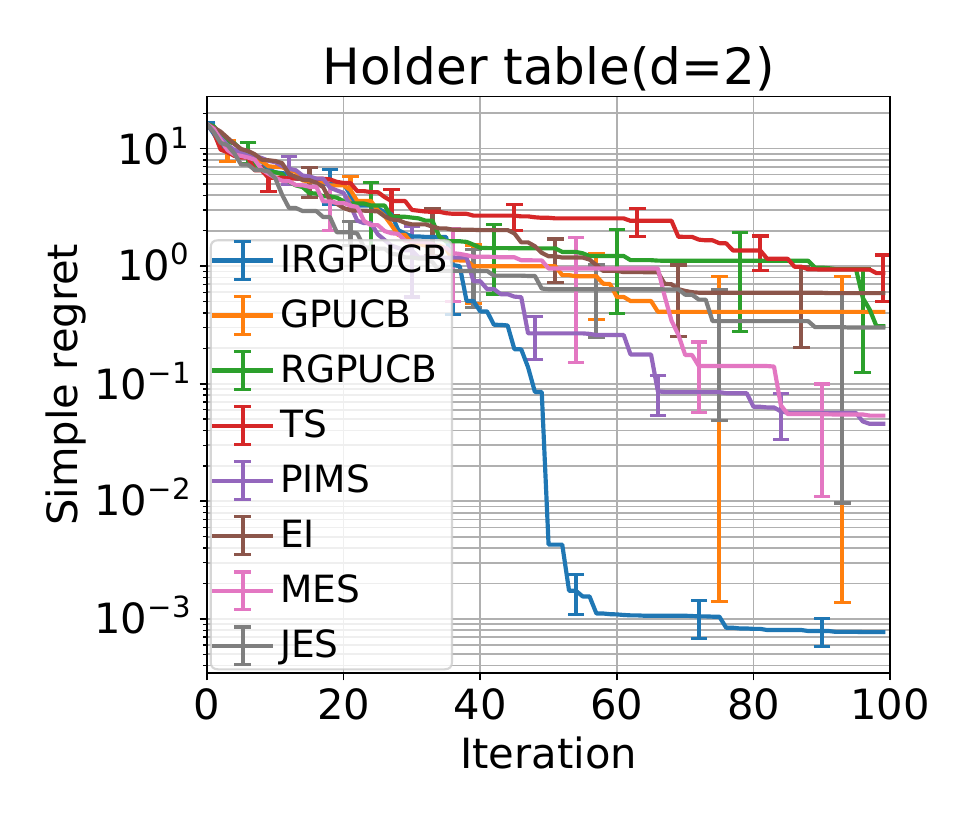}
    \includegraphics[width=0.325\linewidth]{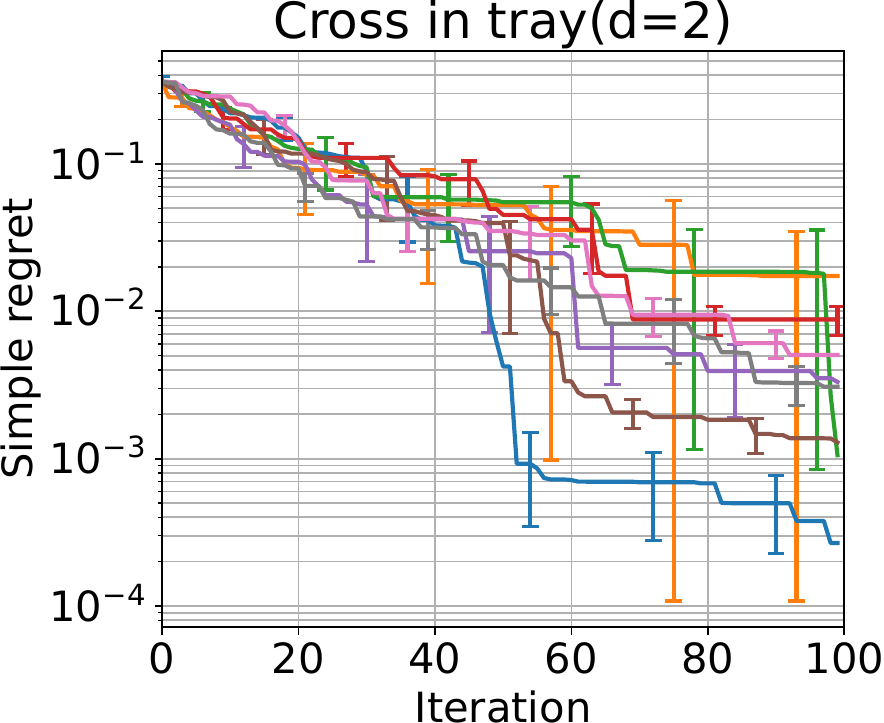}
    \includegraphics[width=0.325\linewidth]{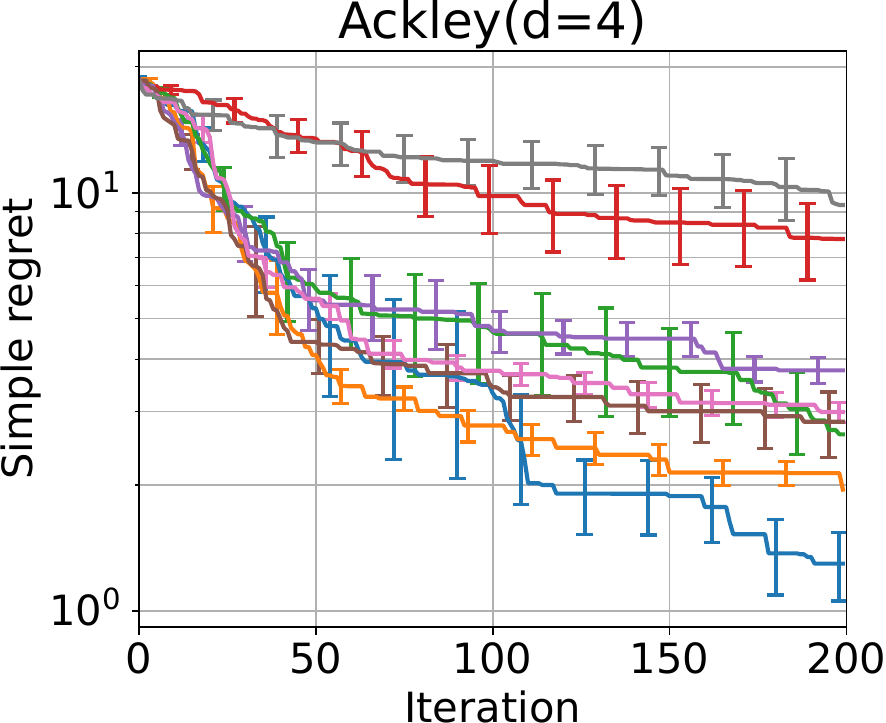}
    \caption{Average and standard errors of simple regret on benchmark functions.}
    \label{fig:benchmark-regret}
    % \label{fig:benchmark-real-regret}
\end{figure}

\subsection{Benchmark Function Experiments}

% \begin{figure*}
%     \centering
%     \begin{subfigure}{\linewidth}
%         \includegraphics[width=0.325\linewidth]{fig/Results_SimMax_Holder_table_log.pdf}
%         \includegraphics[width=0.325\linewidth]{fig/Results_SimMax_Cross_in_tray_log.pdf}
%         % \includegraphics[width=0.325\linewidth]{fig/Results_SimMax_HartMann3_log.pdf}
%         \includegraphics[width=0.325\linewidth]{fig/Results_SimMax_Ackley_log.pdf}
%         \caption{Benchmark functions.}
%         \label{fig:benchmark-regret}
%     \end{subfigure}

%     \begin{subfigure}{\linewidth}
%         \includegraphics[width=0.325\linewidth]{fig/Results_SimMax_Perovskite_log.pdf}
%         \includegraphics[width=0.325\linewidth]{fig/Results_SimMax_P3HT_log.pdf}
%         \includegraphics[width=0.325\linewidth]{fig/Results_SimMax_AgNP_log.pdf}
%         \caption{Materials datasets.}
%         \label{fig:real-regret}
%     \end{subfigure}
    
%     \caption{Average and standard errors of simple regret.}
%     \label{fig:benchmark-real-regret}
% \end{figure*}

We employ three benchmark functions called Holder table ($d=2$), Cross in tray($d=2$), and Ackley ($d=4$) functions, whose analytical forms are shown at \url{https://www.sfu.ca/~ssurjano/optimization.html}.
For each function, we report the average and standard error for $10$ trials using ten random initial datasets $\cD_0$, where $|\cD_0| = 2^d$.
We set the noise variance $\sigma^2 = 10^{-4}$.
We used the Gaussian kernel with automatic relevance determination, whose hyperparameter was selected by marginal likelihood maximization per $5$ iterations \citep{Rasmussen2005-Gaussian}.

In this experiment, since the input domain is continuous, the theoretical choice of the confidence parameter contains an unknown variable.
Thus, we use the heuristic choice for confidence parameters.
For GP-UCB, we set the confidence parameter as $\beta_t = 0.2 d \log(2 t)$, which is the heuristic used in \citep{kandasamy2015-high,Kandasamy2017-Multi}.
For RGP-UCB, we set $\zeta_t \sim {\rm Gamma} (\kappa_t, \theta=1)$ with $\kappa_t = 0.2 d \log(2 t)$ since $\EE[\zeta_t]$ must have the same order as $\beta_t$ (note that $\EE[\zeta_t] = \theta \kappa_t$).
For IRGP-UCB, we set $s = d / 2$ and $\lambda = 1 / 2$.
Note that $\lambda$ equals the theoretical setting, and $s$ captures the dependence on $d$, as shown in Corollary~\ref{theo:BCR_IRGPUCB_continuous}.

Figure~\ref{fig:benchmark-regret} shows the results.
In the Holder table function, JES shows faster convergence until 40 iterations.
However, the regrets of all baseline methods stagnate in $[10^{-1}, 10^0]$.
In contrast, the regret of IRGP-UCB converges to $10^{-3}$ until 60 iterations.
In the Cross in tray function, IRGP-UCB showed rapid convergence at approximately 50 iterations.
In the Ackley function, IRGP-UCB constantly belonged to the top group and showed minimum regret after 125 iterations.
Since the input dimension is relatively large, TS and JES, which depend on the sampled maxima $\*x_*$, deteriorate by over-exploration.
Throughout the experiments, IRGP-UCB outperforms TS, PIMS, GP-UCB, and RGP-UCB, which achieve sub-linear BCR bounds, and EI and MES, which are practically well-used.
This result supports the effectiveness of randomization using the two-parameter exponential distribution.

\begin{figure}[!t]
    \includegraphics[width=0.325\linewidth]{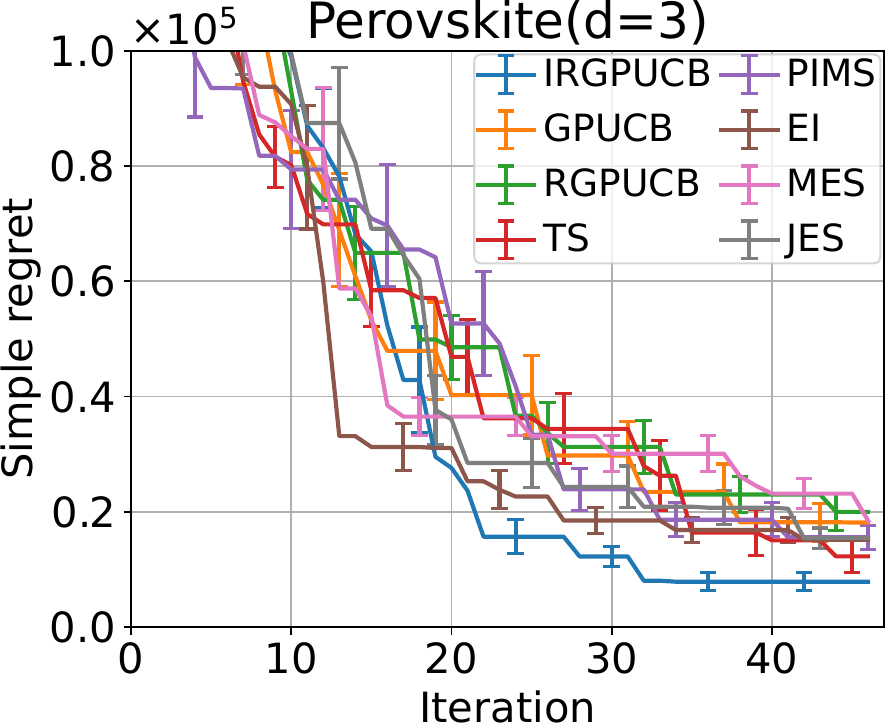}
    \includegraphics[width=0.325\linewidth]{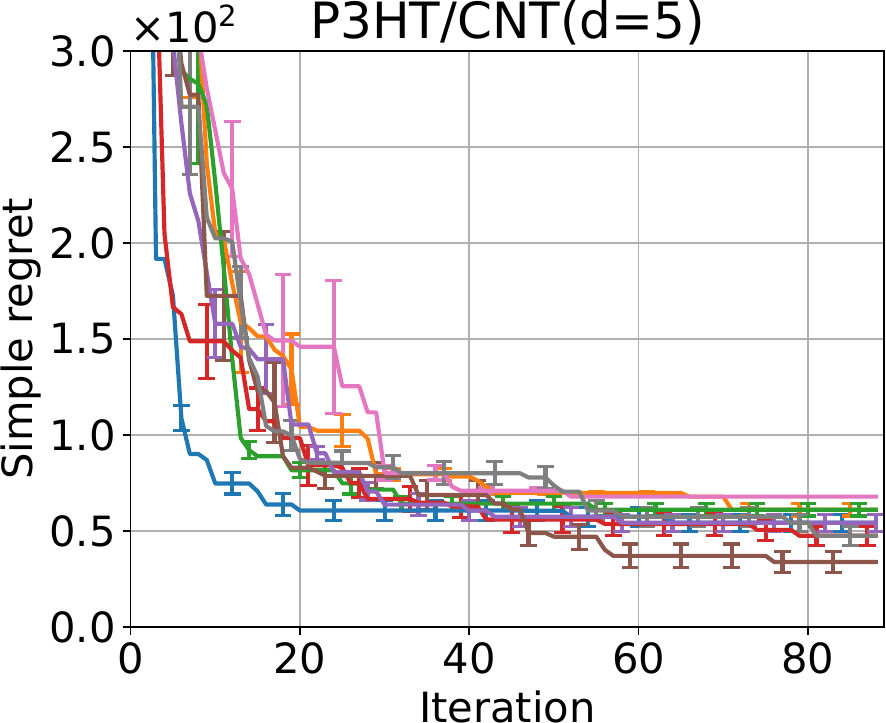}
    \includegraphics[width=0.325\linewidth]{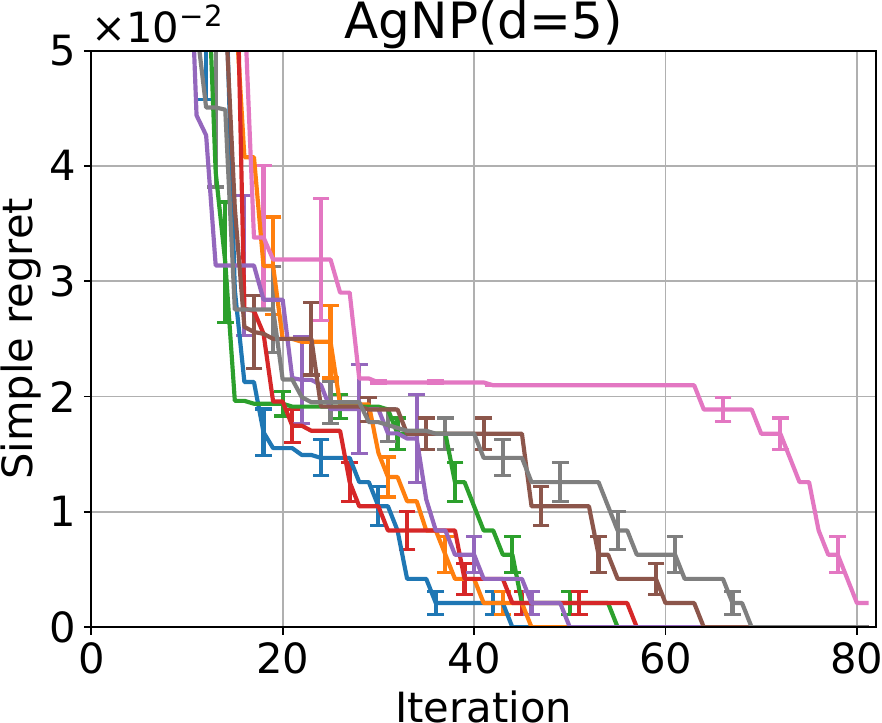}
    \caption{Average and standard errors of simple regret on real-world datasets.}
    \label{fig:real-regret}
    % \label{fig:benchmark-real-regret}
\end{figure}

\subsection{Real-World Dataset Experiments}

This section provides the experimental results on the materials datasets provided in \citep{liang2021benchmarking}.
In the perovskite dataset \citep{sun2021data}, we optimize environmental stability with respect to composition parameters for halide perovskite ($d = 3$ and $|\cX| = 94$).
In the P3HT/CNT dataset \citep{bash2021multi}, we optimize electrical conductivity with respect to composition parameters for the carbon nanotube polymer blend ($d = 5$ and $|\cX| = 178$).
%
% In AutoAM dataset \citep{deneault2021toward}, we optimize printing shapes with the best shape score for the additive manufacturing system with respect to their parameters ($d = 4$ and $|\cX| = 100$).
%
In the AgNP dataset \citep{mekki2021two}, we optimize the absorbance spectrum of synthesized silver nanoparticles with respect to processing parameters for synthesizing triangular nanoprisms ($d = 5$ and $|\cX| = 164$).
See \citep{liang2021benchmarking} for more details about each dataset.

We set the initial dataset size $|\cD_0| = 2$ as with \citep{liang2021benchmarking}.
Since the dataset size is small at earlier iterations and the dataset contains fluctuations from real-world experiments, we observed that hyperparameter tuning could be unstable.
Thus, we optimized the hyperparameters of the RBF kernel in each iteration to avoid repeatedly obtaining samples using an inappropriate hyperparameter.
The other settings matched those used in the benchmark function experiments.

Figure~\ref{fig:real-regret} shows the results.
In the perovskite dataset, IRGP-UCB constantly belonged to the top group and showed the best performance after 20 iterations.
In the P3HT/CNT dataset, EI converged to the smallest value after 60 iterations.
On the other hand, IRGP-UCB shows faster convergence during the first 20 iterations.
In the AgNP dataset, IRGP-UCB found the optimal point after 42 iterations in all the trials.
In this experiment, heuristic methods, EI, MES, and JES, showed worse performance and required at least 60 function evaluations to find the optimal point.
Consequently, we can observe the effectiveness of IRGP-UCB against real-world datasets.

\section{Conclusion}
\label{sec:conculusion}

This paper analyzes the expected regret of the randomized variant of GP-UCB called IRGP-UCB in the Bayesian setting.
Moreover, to fairly compare with the deterministic algorithm, we obtain the high-probability upper bound of the conditional expected regret averaged over the randomness of the GP and the noise, and conditioned on the algorithm's randomness.
In both analyses, if the input domain is finite, increasing the confidence parameter, which can harm the practical efficiency, is avoided.
In addition, by showing the specific instance where GP-UCB using a constant confidence parameter incurs linear expected cumulative regret, we confirmed that randomization plays a key role in avoiding increasing confidence parameters.
Finally, we show the practical effectiveness of the IRGP-UCB through the numerical experiments, including the application to the materials dataset.

Several directions for future work can be considered.
First, whether we can show high-probability regret bounds without increasing the confidence parameter is of interest.
Second, avoiding an increase in the confidence parameter for the continuous domain is also an important direction.
Last, as with the randomized LSE proposed recently \citep{inatsu2024-active}, we may be able to extend IRGP-UCB to various other practical settings, where the usual GP-UCB-based methods have been extended, for example, multi-objective BO \citep{paria2020-flexible,Zuluga2016-epsilon,Suzuki2020-multi,inats2024-bounding}, multi-fidelity BO \citep{Kandasamy2016-Gaussian,Kandasamy2017-Multi,Takeno2020-Multifidelity,Takeno2022-generalized}, parallel BO \citep{Contal2013-Parallel,Desautels2014-Parallelizing}, high-dimensional BO \citep{kandasamy2015-high}, cascade BO \citep{Kusakawa2022-bayesian}, and robust BO \citep{Bogunovic2018-adversarially,iwazaki2021-mean,inatsu2022-bayesian}.

\section*{Acknowledgements}
This work was supported by 
JSPS/MEXT KAKENHI Grant Number (JP20H00601, JP21H03498, 22H00300, JP21J14673, JP23K16943, JP23K19967, and JP24K20847),
MEXT Program: Data Creation and Utilization-Type Material Research and Development Project Grant Number JPMXP1122712807, 
JST ACT-X Grant Number (JPMJAX23CD and JPMJAX24C3),
JST PRESTO Grant Number JPMJPR24J6,
and RIKEN Center for Advanced Intelligence Project.

%%%%%%%%%%%%%%%%%%%%%%%%%%%%%%%%%%%%%%%%%%%%%%%%%%%%%%%%%%%%%%%%%%%%%%%%%%%%%%%%%%%%%%%
% % \subsubsection*{References}
\bibliography{ref}
\bibliographystyle{abbrvnat}

% %%%%%%%%%%%%%%%%%%%%%%%%%%%%%%%%%%%%%%%%%%%%%%%%%%%%%%%%%%%%%%%%%%%%%%%%%%%%%%%%%%%%%%
\clearpage
\appendix
% \onecolumn
% \thispagestyle{empty}
% \begin{center}
%  {\Large {\bf
%  Supplementary Materials for the Submission:\\
%  ``Randomized Gaussian Process Upper Confidence Bound \\ with Regret Bounds''
%  }
%  }
% \end{center}

% \input{manuscripts/9_appendix_BSR_with_recommendation}
% \input{manuscripts/9_appendix_BCR_GPUCB}
% \input{manuscripts/9_appendix_BCR_RGPUCB}
\section{Proofs for Expected Regret Bounds of IRGP-UCB}
%%%%%%%%%%%%%%%%%%%%%%%%%%%%%%%%%%%%%%%%%%%%%%%%%%%%%%%%%%%%%%%%%%%%%%%%%%%%%%%%
% \section{Proofs of Lemmas~\ref{lem:bound_srinivas} and \ref{lem:bound_RGPUCB}}

\subsection{Proof of Lemma~\ref{lem:bound_srinivas}}
\label{app:proof_srinivas}
The main difference between Lemma~\ref{lem:bound_srinivas} and \citep[Lemma 5.1 in ][]{Srinivas2010-Gaussian} is that Lemma~\ref{lem:bound_srinivas} does not consider the union bound for all $t \geq 1$. 
That is, Lemma 5.1 in \citet{Srinivas2010-Gaussian} considers the following probability:
\begin{align*}
    \Pr \left( f(\*x) \leq \mu_{t-1}(\*x) + \beta^{1/2}_t \sigma_{t-1}(\*x), \forall \*x \in \cX, \forall t \geq 1 \right).
\end{align*}
On the other hand, Lemma~\ref{lem:bound_srinivas} bounds the following probability with one fixed $t$:
\begin{align*}
    \myPr_t \left( f(\*x) \leq \mu_{t-1}(\*x) + \beta^{1/2}_{\delta} \sigma_{t-1}(\*x), \forall \*x \in \cX\right),
\end{align*}
where $\beta^{1/2}_{\delta}$ does not depend on $t$ as described below.

Assume the same condition as in Lemma~\ref{lem:bound_srinivas}, and thus, $\beta_\delta = 2\log\bigl(|\cX| / (2\delta)\bigr)$. 
Then, for all $\*x \in \cX$ and $\cD_{t-1}$, we see that
\begin{align*}
    \myPr_t \left( f(\*x) > \mu_{t-1}(\*x) + \beta^{1/2}_{\delta} \sigma_{t-1}(\*x) \right) \leq \frac{\delta}{|\cX|},
\end{align*}
where we use an elementary result of the Gaussian distribution shown in Lemma~\ref{lem:Gauss_tail_bound} in Appendix~\ref{app:lemmas}. 
Then, we can obtain the following bound:
\begin{align*}
    \myPr_t \left( f(\*x) > \mu_{t-1}(\*x) + \beta^{1/2}_{\delta} \sigma_{t-1}(\*x), \exists \*x \in \cX \right)
    &\leq \sum_{\*x \in \cX} \myPr_t \left( f(\*x) > \mu_{t-1}(\*x) + \beta^{1/2}_{\delta} \sigma_{t-1}(\*x) \right) \\
    &\leq \delta.
\end{align*}
Therefore, 
\begin{align*}
    &\myPr_t \left( f(\*x) \leq \mu_{t-1}(\*x) + \beta^{1/2}_{\delta} \sigma_{t-1}(\*x), \forall \*x \in \cX \right) \\
    &= 1 - \myPr_t \left( f(\*x) > \mu_{t-1}(\*x) + \beta^{1/2}_{\delta} \sigma_{t-1}(\*x), \exists \*x \in \cX \right) \geq 1 - \delta,
\end{align*}
which concludes the proof.

%%%%%%%%%%%%%%%%%%%%%%%%%%%%%%%%%%%%%%%%%%%%%%%%%%%%%%%%%%%%%%%%%%%%%%%%%%%%%%%%
\subsection{Proof of Lemma~\ref{lem:bound_RGPUCB}}
\label{app:proof_lemma_IRGPUCB}

This section provides detailed proof of the following Lemma~\ref{lem:bound_RGPUCB}:
\begin{replemma}{lem:bound_RGPUCB}
    Let $f \sim \cG \cP (0, k)$, where $k$ is a positive semidefinite kernel, and $\cX$ be finite.
    Assume that $\zeta_t$ follows a shifted exponential distribution with $s_t = 2 \log (|\cX| / 2)$ and $\lambda = 1/2$.
    Then, for any given $\cD_{t-1}$, the following inequality holds:
    \begin{align*}
        \EE_t[f(\*x^*)] \leq \EE_t \left[\max_{\*x \in \cX} \mu_{t-1}(\*x) + \zeta^{1/2}_t \sigma_{t-1}(\*x) \right],
    \end{align*}
    for all $t \geq 1$.
\end{replemma}
\begin{proof}
    Fix the dataset $\cD_{t-1}$ and $\delta \in (0, 1)$.
    From Lemma~\ref{lem:bound_srinivas}, we can see that
    \begin{align*}
        \myPr_t \left( f(\*x^*) \leq \max_{\*x \in \cX} \mu_{t-1}(\*x) + \beta^{1/2}_{\delta} \sigma_{t-1}(\*x) \right) \geq 1 - \delta,
    \end{align*}
    where $\beta_{\delta} =  2 \log (|\cX| / (2\delta))$.
    Note that the probability is taken only with $f(\*x^*)$ since $\cD_{t-1}$ is fixed.
    Using the cumulative distribution function $F_t(\cdot) \coloneqq \myPr_t(f(\*x^*) \leq \cdot)$ and its inverse function $F^{-1}_t$, we can rewrite
    \begin{align*}
        F_t \left( \max_{\*x \in \cX} \mu_{t-1}(\*x) + \beta^{1/2}_{\delta} \sigma_{t-1}(\*x) \right) &\geq 1 - \delta \\ 
        \Longleftrightarrow \max_{\*x \in \cX} \mu_{t-1}(\*x) + \beta^{1/2}_{\delta} \sigma_{t-1}(\*x) &\geq F^{-1}_t ( 1 - \delta ),
    \end{align*}
    since $F^{-1}_t$ is a monotone increasing function.
    Since $\delta \in (0, 1)$ is arbitrary, we can see that
    \begin{align*}
         \Pr \biggl( \max_{\*x \in \cX} \mu_{t-1}(\*x) + \beta^{1/2}_{U} \sigma_{t-1}(\*x) \geq F^{-1}_t ( 1 - U ) \biggr) = 1,
    \end{align*}
    where $U \sim {\rm Uni}(0, 1)$ and probability is taken by the randomness of $U$.
    Then, because of the basic property of the expectation $\EE[X] \leq \EE[Y]$ if $\Pr(X \leq Y) = 1$, we can obtain,
    \begin{align*}
         \EE_{U} \left[ \max_{\*x \in \cX} \mu_{t-1}(\*x) + \beta^{1/2}_{U} \sigma_{t-1}(\*x)   \right]
         &\geq \EE_{U} \bigl[ F^{-1}_t ( 1 - U )   \bigr] \\
         &= \EE_{U} \bigl[ F^{-1}_t ( U )   \bigr],
    \end{align*}
    where we use the fact that $1 - U$ also follows ${\rm Uni}(0, 1)$.
    Since $F^{-1}_t ( U )$ and $f(\*x^*)$ are identically distributed as with the inverse transform sampling, we obtain
    \begin{align*}
         \EE_{U} \left[ \max_{\*x \in \cX} \mu_{t-1}(\*x) + \beta^{1/2}_{U} \sigma_{t-1}(\*x)   \right]
         \geq \EE_t \bigl[ f(\*x^*) \bigr].
    \end{align*}
    Then, $\beta_U$ can be decomposed as follows:
    \begin{align*}
        \beta_U = 2 \log (|\cX| / 2) - 2 \log(U).
    \end{align*}
    From the inverse transform sampling, $ - 2 \log(U) \sim {\rm Exp} (\lambda = 1/2)$.
    Hence, $\beta_U$ follows a shifted exponential distribution with $s = 2 \log (|\cX| / 2)$ and $\lambda = 1/2$.
    Therefore, we obtain the following:
    \begin{align*}
         \EE_t \left[ \max_{\*x \in \cX} \mu_{t-1}(\*x) + \zeta^{1/2} \sigma_{t-1}(\*x) \right]
         \geq \EE_t \bigl[ f(\*x^*)  \bigr].
    \end{align*}
\end{proof}

%%%%%%%%%%%%%%%%%%%%%%%%%%%%%%%%%%%%%%%%%%%%%%%%%%%%%%%%%%%%%%%%%%%%%%%%%%%%%%%%
\subsection{Proof of Theorem~\ref{theo:BCR_IRGPUCB_discrete}}
\label{app:proof_discrete}

Using Lemma~\ref{lem:bound_RGPUCB}, we can obtain the proof of the following theorem for finite input domains:
\begin{reptheorem}{theo:BCR_IRGPUCB_discrete}
    Let $f \sim \cG \cP (0, k)$, where $k$ is a positive semidefinite kernel and $k(\*x, \*x) \leq 1$, and $\cX$ be finite.
    Assume that $\zeta_t$ follows a shifted exponential distribution with $s_t = 2 \log (|\cX| / 2)$ and $\lambda = 1/2$ for any $t \geq 1$.
    Then, by running IRGP-UCB with $\zeta_t$, BCR can be bounded as follows: 
    \begin{align*}
        {\rm BCR}_T \leq \sqrt{C_1 C_2 T \gamma_T},
    \end{align*}
    where $C_1 \coloneqq 2 / \log(1 + \sigma^{-2})$ and $C_2 \coloneqq 2 + 2 \log (|\cX| / 2)$. 
\end{reptheorem}
\begin{proof}
    From Lemma~\ref{lem:bound_RGPUCB}, we obtain
    \begin{align}
        {\rm BCR}_T
        &= \sum_{t=1}^T \EE[f(\*x^*) - f(\*x_t)] \\
        &= \sum_{t=1}^T \EE[f(\*x^*) - v_t(\*x_t) + v_t(\*x_t) - f(\*x_t)] \\
        &\leq \sum_{t=1}^T \EE[v_t(\*x_t) - f(\*x_t)],
    \end{align}
    where $v_t(\*x) = \mu_{t-1}(\*x) + \zeta_t^{1/2}(\*x)$.
    Then, we can easily derive the bound as follows:
    \begin{align}
        {\rm BCR}_T
        &\leq \sum_{t=1}^T \EE[v_t(\*x_t) - f(\*x_t)] \\
        &= \sum_{t=1}^T \EE_{\cD_{t-1}, \zeta_t} [ \EE [v_t(\*x_t) - f(\*x_t) | \cD_{t-1}, \zeta_t]] \\
        &= \sum_{t=1}^T \EE_{\cD_{t-1}} \left[ \EE_{\zeta_t} [v_t(\*x_t)  - \mu_{t-1}(\*x_t)] \right] \\
        &= \sum_{t=1}^T \EE_{\cD_{t-1}, \zeta_t} [ \zeta_t^{1/2} \sigma_{t-1}(\*x_t) ] \\
        &= \EE \left[ \sum_{t=1}^T \zeta_t^{1/2} \sigma_{t-1}(\*x_t) \right] \\
        &\leq \EE \left[\sqrt{ \sum_{t=1}^T \zeta_t \sum_{t=1}^T \sigma_{t-1}^2 (\*x_t) } \right] && \bigl( \because \text{Cauchy--Schwarz  inequality} \bigr) \\
        &\leq \EE \left[\sqrt{ \sum_{t=1}^T \zeta_t } \right] \sqrt{C_1 \gamma_T} && \bigl( \because \text{Lemma~5.4 in \citep{Srinivas2010-Gaussian}} \bigr) \\
        &\leq \sqrt{ \sum_{t=1}^T \EE [ \zeta_t ]} \sqrt{C_1 \gamma_T} && \bigl( \because \text{Jensen's inequality} \bigr) \\
        &= \sqrt{C_1 C_2 T \gamma_T}, && \bigl( \because \text{Definition of $\zeta_t$} \bigr)
    \end{align}
    where $C_2 = \EE [ \zeta_t ] = 2 + 2\log(|\cX|/2)$.
    This concludes the proof.
\end{proof}

%%%%%%%%%%%%%%%%%%%%%%%%%%%%%%%%%%%%%%%%%%%%%%%%%%%%%%%%%%%%%%%%%%%%%%%%%%%%%%%%
\subsection{Proof of Theorem~\ref{theo:BCR_IRGPUCB_continuous}}
\label{app:proof_continuous}

\begin{reptheorem}{theo:BCR_IRGPUCB_continuous}
    Let $f \sim \cG \cP (0, k)$, where $k$ is a positive semidefinite kernel, $k(\*x, \*x) \leq 1$, and Assumption~\ref{assump:continuous_X} holds.
    Assume that $\zeta_t$ follows a shifted exponential distribution with $s_t = 2d \log(bdr t^2 \bigl( \sqrt{\log (ad)} + \sqrt{\pi} / 2\bigr)) - 2 \log 2$ and $\lambda = 1/2$ for any $t \geq 1$.
    Then, by running IRGP-UCB, BCR can be bounded as follows: 
    \begin{align*}
        {\rm BCR}_T \leq \frac{\pi^2}{6} + \sqrt{C_1 T \gamma_T (2 + s_T)},
    \end{align*}
    where $C_1 \coloneqq 2 / \log(1 + \sigma^{-2})$. 
\end{reptheorem}
\begin{proof}
    % Purely for the sake of analysis, we used a set of discretization $\cX_t \subset \cX$ for $t \geq 1$.
    %
    % For any $t \geq 1$, let $\cX_t \subset \cX$ be a finite set with each dimension equally divided into $\tau_t = bdr t^2 \bigl(\sqrt{\log (ad)} + \sqrt{\pi} / 2 \bigr)$.
    We consider a finite discretized input set $\cX_t \subset \cX$ for $t \geq 1$ for theoretical analysis, though $\cX_t$ is not related to the actual algorithm.
    Let $\cX_t = \{\frac{r}{\tau_t + 1}, \dots, \frac{\tau_t r}{\tau_t + 1} \}^d \subset \cX$, where each coordinate takes values from a uniform grid of $\tau_t = bdr u_t \bigl( \sqrt{\log (ad)} + \sqrt{\pi} / 2 \bigr)$ points.
    Thus, $|\cX_t| = \tau_t^d$.
    In addition, we define $[\*x]_t$ as the nearest point in $\cX_t$ of $\*x \in \cX$.

    Then, we decompose BCR as follows:
    \begin{align}
        {\rm BCR}_T
        &= \sum_{t=1}^T \EE \Biggl[ f(\*x^*) - f( [\*x^*]_t ) + f( [\*x^*]_t ) - \max_{\*x \in \cX_t} v_t(\*x) + \max_{\*x \in \cX_t} v_t(\*x) - v_t(\*x_t) + v_t(\*x_t) - f(\*x_t) \Biggl].
    \end{align}
    Obviously, $\EE\bigl[\max_{\*x \in \cX_t} v_t(\*x) - v_t(\*x_t) \bigr] \leq 0$ from the definition of $\*x_t$.
    Furthermore, we observe the following:
    \begin{align*}
        \EE \left[ f( [\*x^*]_t ) - \max_{\*x \in \cX_t} v_t(\*x) \right] \leq \EE \left[ \max_{\*x \in \cX_t} f( \*x ) - \max_{\*x \in \cX_t} v_t(\*x) \right],
    \end{align*}
    which can be bounded above by $0$ by setting $s_t = 2\log(|\cX_t| / 2)$ and $\lambda = 1 / 2$ from Lemma~\ref{lem:bound_RGPUCB}.
    Therefore, we obtain the following:
    \begin{align}
        {\rm BCR}_T
        &\leq \sum_{t=1}^T \EE \left[ f(\*x^*) - f( [\*x^*]_t ) + v_t(\*x_t) - f(\*x_t) \right] \\
        &= \sum_{t=1}^T \EE \left[ f(\*x^*) - f( [\*x^*]_t ) \right] + \sum_{t=1}^T \EE \left[ v_t(\*x_t) - f(\*x_t) \right]. \label{eq:BCR_continuous_two_term_bound}
    \end{align}

    First, we consider the first term $ \sum_{t=1}^T \EE \bigl[ f(\*x^*) - f( [\*x^*]_t ) \bigr]$.
    From Lemma~\ref{lem:discretized_error}, we can obtain the following:
    \begin{align}
        \sum_{t=1}^T \EE \left[ f(\*x^*) - f( [\*x^*]_t ) \right] 
        &\leq \sum_{t=1}^T \EE \left[ \sup_{\*x \in \cX} f(\*x) - f( [\*x]_t ) \right] \\
        &\leq \sum_{t=1}^T \frac{1}{t^2} && \bigl( \because \text{ Lemma~\ref{lem:discretized_error} } \bigr) \\
        &\leq \frac{\pi^2}{6}, && \left(\because \sum_{t=1}^\infty \frac{1}{t^2} = \frac{\pi^2}{6} \right) \label{eq:discrete_points_distance_bound}
    \end{align}
    where $u_t$ in Lemma~\ref{lem:discretized_error} corresponds to $t^2$.

    The second term is bounded as follows:
    \begin{align}
        \sum_{t=1}^T \EE \left[ v_t(\*x_t) - f(\*x_t) \right]
        &\leq \sqrt{ \sum_{t=1}^T \EE[\zeta_t] C_1 \gamma_T }, && \bigl(\because \text{See the proof of Theorem~\ref{theo:BCR_IRGPUCB_discrete}} \bigr) \label{eq:continouos_bound_width_bound}
    \end{align}
    where 
    \begin{align}
        \EE[\zeta_t]
        &= 2 + 2\log(|\cX_t| / 2) \\
        &= 2 + 2d \log(bdr t^2 \bigl( \sqrt{\log (ad)} + \sqrt{\pi} / 2\bigr)) - 2 \log 2, 
    \end{align}
    which is monotone increasing.
    Therefore, we obtain the following:
    \begin{align}
        \sum_{t=1}^T \EE \left[ v_t(\*x_t) - f(\*x_t) \right]
        &\leq \sqrt{ C_1 T \gamma_T \EE[\zeta_T] },
    \end{align}
    where $\EE[\zeta_T] = 2 + s_T = 2 + 2d \log(bdr T^2 \bigl( \sqrt{\log (ad)} + \sqrt{\pi} / 2\bigr)) - 2 \log 2$.
    Consequently, combining Eqs.~\eqref{eq:BCR_continuous_two_term_bound}, \eqref{eq:discrete_points_distance_bound}, and \eqref{eq:continouos_bound_width_bound} concludes the proof.
\end{proof}

\section{Proofs for Conditional Expected Regret Bounds of IRGP-UCB}
\label{app:proof_conditional_hpcr}

%%%%%%%%%%%%%%%%%%%%%%%%%%%%%%%%%%%%%%%%%%%%%%%%%%%%%%%%%%%%%%%%%%%%%%%%%%%%%%%%
\subsection{Proof of Theorem~\ref{theo:HPCER_IRGPUCB_discrete}}
\label{app:proof_conditional_hpcr_discrete}
%%%%%%%%%%%%%%%%%%%%%%%%%%%%%%%%%%%%%%%%%%%%%%%%%%%%%%%%%%%%%%%%%%%%%%%%%%%%%%%%

For the proof of Theorem~\ref{theo:HPCER_IRGPUCB_discrete}, we use the properties of subgaussian random variables.
Therefore, before the proof of Theorem~\ref{theo:HPCER_IRGPUCB_discrete}, we provide several propositions.

The definition of the subgaussian random variable is shown below \citep[Definition~3.5 of][]{Handel-HDP}:
\begin{definition}
    A random variable is called $\sigma^2$-subgaussian if the following is satisfied:
    \begin{align*}
        \EE[\exp(\lambda X)] \leq \exp(\sigma^2 \lambda^2 / 2),
    \end{align*}
    for all $\lambda \in \RR$.
    Then, $\sigma^2$ is called the variance proxy.
    \label{def:subgaussian}
\end{definition}
Several equivalent definitions are known \citep[Proposition 2.5.2 of][]{Vershynin2018-high}.
In particular, we use the following relations:
\begin{proposition}
    Assume that the random variable $X$ satisfies the following property:
    \begin{align*}
        \Pr(|X| > c) \leq 2 \exp(-c^2 / A^2),
    \end{align*}
    for all $c \geq 0$ and $\EE[X] = 0$.
    Then, $X$ is $9A^2$-subgaussian.
    \label{prop:tail_prob_to_subgaussian}
\end{proposition}
\begin{proof}
    For example, by combining Problems~3.1 (d) and (e) of \citet{Handel-HDP}, we can obtain this result.
\end{proof}
\begin{proposition}
    Let $X$ be $\sigma^2$-subgaussian.
    Then, 
    \begin{align*}
        \Pr(X > c) \leq \exp(-c^2 / 2 \sigma^2),
    \end{align*}
    and
    \begin{align*}
        \Pr(|X| > c) \leq 2 \exp(-c^2 / 2 \sigma^2).
    \end{align*}
    \label{prop:subgaussian_to_tail_prob}
\end{proposition}
\begin{proof}
    See, for example, Problem~3.1 (c) of \citet{Handel-HDP} or the proof for (v) $\Rightarrow$ (i) of Proposition 2.5.2 in \citep{Vershynin2018-high}.
\end{proof}

Furthermore, we use the following form of Azuma's inequality \citep[Lemma~3.7 of][]{Handel-HDP,azuma1967weighted}:
\begin{lemma}[Azuma's inequality]
    Let $\{ \cF_t \}_{t \leq T}$ be any filtration, and let $X_1, \dots, X_T$ be random variables that satisfy the following properties for $t = 1, \dots, T$:
    \begin{enumerate}
        \item Martingale difference property: $X_t$ is $\cF_t$-measurable and $\EE[X_t \mid \cF_{t-1}] = 0$.
        \item Conditional subgaussian property: $\EE[\exp(\lambda X_t) \mid \cF_{t-1}] \leq \exp(\sigma^2_t \lambda^2 / 2)$ for all $\lambda \in \RR$ almost surely.
    \end{enumerate}
    Then, the sum $\sum_{t=1}^T X_t$ is subgaussian with variance proxy $\sum_{t=1}^T \sigma^2_t$.
    \label{lem:azuma}
\end{lemma}

To show the subgaussian property the variable $A_1$ in the following proof of Theorem~\ref{theo:HPCER_IRGPUCB_discrete}, we use the following proposition shown in, for example, Theorem~5.2.2 of \citet{Vershynin2018-high}, Proposition 4 of \citet{zeitouni2014gaussian}, and Lemma 6.4 of \citet{biskup2020extrema}:
\begin{proposition}[Concentration for Lipshitz functions]
    Let $\*X = (X_1, \dots, X_k)$ be a random vector follows $\cN(\*0, \*I_k)$, where $\*I_k$ is $k$-dimensional identity matrix.
    Suppose that $h: \RR^k \rightarrow \RR$ is Lipshitz; that is, there exists a constant $L_h \in \RR$ such that
    \begin{align*}
        \| h(\*a) - h(\*b) \|_2 \leq L_h \| \*a - \*b \|_2,
    \end{align*}
    for all $\*a, \*b \in \RR^k$, where $\|\cdot\|_2$ is L2 norm. Then,
    \begin{align*}
        \Pr \left( \left| h(\*X) - \EE[h(\*X)] \right| > c \right) \leq 2 \exp \left\{ - \frac{c^2}{2 L_h^2} \right\}.
    \end{align*}
    \label{prop:lipshitz_contraction}
\end{proposition}

%%%%%%%%%%%%%%%%%%%%%%%%%%%%%%%%%%%%%%%%%%%%%%%%%%%%%%%%%%%%%%%%%%%%
Then, we show the following theorem:
\begin{reptheorem}{theo:HPCER_IRGPUCB_discrete}
    Let $f \sim \cG \cP (0, k)$, where $k$ is a positive semidefinite kernel and $k(\*x, \*x) \leq 1$, and $\cX$ be finite.
    Assume that $\zeta_t$ follows a shifted exponential distribution with $s_t = 2 \log (|\cX| / 2)$ and $\lambda = 1/2$ for any $t \geq 1$.
    Then, by running IRGP-UCB with $\zeta_t$, the conditional expected regret can be bounded with probability at least $1 - \delta$ as follows: 
    \begin{align*}
        % &\Pr \left( \EE_{f, \{\epsilon_t\}_{t \geq 1}} \left[ \sum_{t=1}^T f(\*x^*) - f(\*x_t) \mid \{\zeta_t\}_{t \geq 1} \right] \leq U(T, \delta) \right) \geq 1 - \delta, \forall T \geq 1, \\
        %
        &\Pr \left( \forall T \geq 1, \EE_{f, \{\epsilon_t\}_{t \geq 1}} \left[ \sum_{t=1}^T f(\*x^*) - f(\*x_t) \mid \{\zeta_t\}_{t \geq 1} \right] \leq U \bigl(T, \delta \bigr) \right) \geq 1 - \delta, 
    \end{align*}
    where 
    \begin{align*}
        U(T, \delta) 
        &= 6\sqrt{T \log\left(\frac{\pi^2 T^2}{3\delta} \right)} + \sqrt{C_1 \gamma_T \left( T s_T + T + 2\sqrt{T \log\left(\frac{\pi^2 T^2}{3\delta} \right)} + 2\log\left(\frac{\pi^2 T^2}{3\delta} \right)\right) } \\
        % &= O \left( \sqrt{T \gamma_T \log|\cX| + \gamma_T \sqrt{T \log(1/\delta)}} + \gamma_T \log(1 / \delta) \right),
    \end{align*}
    and $C_1 \coloneqq 2 / \log(1 + \sigma^{-2})$.
\end{reptheorem}
\begin{proof}
    First, we show an upper bound for some fixed $T \geq 1$.
    For any $\{\zeta_t\}_{t \geq 1}$, we obtain
    \begin{align*}
        & \sum_{t=1}^T \EE_{f, \{\epsilon_t\}_{t \geq 1}} \left[  f(\*x^*) - f(\*x_t) \mid \{\zeta_t\}_{t \geq 1} \right] \\
        &= \sum_{t=1}^T \EE_{ \cD_{t-1}} \left[  \EE_{f \mid \cD_{t-1}} \left[  f(\*x^*) \right] -  \mu_{t-1}(\*x_t) \mid \{\zeta_{i} \}_{i \leq t} \right] \\
        &= \sum_{t=1}^T \EE_{ \cD_{t-1}} \left[ 
                \EE_{f \mid \cD_{t-1}} \left[  f(\*x^*) \right] 
                - \EE_{\zeta_t \mid \cD_{t-1}} \left[ v_t(\*x_t) \right] + \EE_{\zeta_t \mid \cD_{t-1}} \left[ v_t(\*x_t) \right]
                - v_t(\*x_t) + v_t(\*x_t)
                -  \mu_{t-1}(\*x_t) 
            \mid \{\zeta_{i} \}_{i \leq t} \right] \\
        &\leq \sum_{t=1}^T \EE_{ \cD_{t-1}} \left[ 
                \EE_{\zeta_t \mid \cD_{t-1}} \left[ v_t(\*x_t) \right]
                - v_t(\*x_t) + \zeta_t^{1/2} \sigma_{t-1}(\*x_t)
            \mid \{\zeta_{i} \}_{i \leq t} \right] \qquad (\because \text{Lemma~\ref{lem:bound_RGPUCB}}) \\
        &=
            \underbrace{\sum_{t=1}^T \biggl\{ \EE_{ \cD_{t-1}, \zeta_t} \left[ v_t(\*x_t) \mid \{\zeta_{i} \}_{i < t} \right]
            - \EE_{ \cD_{t-1}} \left[ v_t(\*x_t) \mid \{\zeta_{i} \}_{i \leq t} \right] \biggr\}}_{A_1}
            + \underbrace{\EE \left[ \sum_{t=1}^T \zeta_t^{1/2} \sigma_{t-1}(\*x_t) \mid \{\zeta_{t} \}_{t > 1} \right]}_{A_2},
    \end{align*}
    where $v_t(\*x_t) = \mu_{t-1}(\*x_t) + \zeta_t^{1/2} \sigma_{t-1}(\*x_t)$.

    First, regarding $A_2$, the following inequality holds with probability at least $1 - \delta_2$,
    \begin{align*}
        &\EE \left[ \sum_{t=1}^T \zeta_t^{1/2} \sigma_{t-1}(\*x_t) \mid \{\zeta_{t} \}_{t > 1} \right] \\
        &\leq \sqrt{\sum_{t=1}^T \zeta_t} \EE \left[ \sqrt{\sum_{t=1}^T \sigma_{t-1}^2(\*x_t)} \mid \{\zeta_{t} \}_{t > 1} \right] && (\because \text{Cauchy--Schwartz inequality}) \\
        &\leq \sqrt{C_1 \gamma_T \sum_{t=1}^T \zeta_t} && (\because \text{Lemma 5.4 of \citet{Srinivas2010-Gaussian}}) \\
        &\leq \sqrt{C_1 \gamma_T \left( T s_T + T + 2\sqrt{T \log(1 / \delta_2)} + 2\log(1 / \delta_2)\right) }, && (\because \text{Lemma~\ref{lem:Laurent}})
    \end{align*}
    where the last inequality holds with high-probability since $\sum_{t=1}^T \{ \zeta_t - s_t \}$ follows a chi-square distribution with $T$ degrees of freedom.
    Note that, regarding the first inequality, $\sqrt{\sum_{t=1}^T \zeta_t}$ can be put outside of the conditional expectation since the expectation is taken with $f$ and $\{ \epsilon_t \}_{t \geq 1}$.

    Second, we consider $A_1$.
    We will show that $A_1$ is an $18T$-subgaussian random variable.
    Let history $\cH_t = \{ \zeta_i \}_{i \leq t}$.
    Then, we can see that
    \begin{align*}
        &\EE_{ \cD_{t-1}, \zeta_t} \left[ v_t(\*x_t) \mid \{\zeta_{i} \}_{i < t} \right]
            - \EE_{ \cD_{t-1}} \left[ v_t(\*x_t) \mid \{\zeta_{i} \}_{i \leq t} \right] \\
        &= \EE_{ \cD_{t-1}, \zeta_t} \left[ v_t(\*x_t) \mid \cH_{t-1} \right]
        - \EE_{ \cD_{t-1}} \left[ v_t(\*x_t) \mid \cH_t \right],
    \end{align*}
    and, for any $\cH_{t-1}$, the conditional expectation is zero as follows:
    \begin{align*}
        &\EE_{\zeta_t} \left[ \EE_{ \cD_{t-1}, \zeta_t} \left[ v_t(\*x_t) \mid \cH_{t-1} \right]
        - \EE_{ \cD_{t-1}} \left[ v_t(\*x_t) \mid \cH_t \right] \mid \cH_{t-1}\right] \\
        &= \EE_{ \cD_{t-1}, \zeta_t} \left[ v_t(\*x_t) \mid \cH_{t-1} \right]
        - \EE_{ \cD_{t-1}, \zeta_t} \left[ v_t(\*x_t) \mid \cH_{t-1} \right]
        = 0.
    \end{align*}
    In addition, from the definition of measurability, $\EE_{ \cD_{t-1}} \left[ v_t(\*x_t) \mid \cH_t \right]$ is $\cH_t$-mesuarable.

    Then, we derive the conditional subgaussian property of $\EE_{ \cD_{t-1}, \zeta_t} \left[ v_t(\*x_t) \mid \cH_{t-1} \right] - \EE_{ \cD_{t-1}} \left[ v_t(\*x_t) \mid \cH_t \right]$.
    Let a function $h: \RR^2 \rightarrow \RR$ be
    \begin{align*}
        h(\*a) = \EE_{\cD_{t-1}} \left[ \max_{\*x \in \cX} \left\{ \mu_{t-1}(\*x) + \sqrt{ 2 \log(|\cX| / 2) + \| \*a \|_2^2 } \sigma_{t-1}(\*x) \right\} \mid \cH_{t-1} \right].
    \end{align*}
    Note that $\bigl(h(\*X) \mid \cH_{t-1}\bigr) \overset{\mathrm{d}}{=} \bigl(\EE_{\cD_{t-1}} \left[ v_t(\*x_t) \mid \zeta_t \right] \mid \cH_{t-1} \bigr)$ for any fixed $\cH_{t-1}$, where $\*X \sim \cN(\*0, \*I_2)$ and $\overset{\mathrm{d}}{=}$ means the equivalence of the distribution, since $\| \*X \|_2^2$ follows the exponential distribution with $\lambda = 1/2$.
    Then, for any fixed $\cH_{t-1}$, $h$ is $1$-Lipschitz function as shown below:
    \begin{align*}
        &|h(\*a) - h(\*b)| \\
        &= \bigg| \EE_{\cD_{t-1}} \biggl[ \max_{\*x \in \cX} \left\{ \mu_{t-1}(\*x) + \sqrt{ 2 \log(|\cX| / 2) + \| \*a \|_2^2 } \sigma_{t-1}(\*x) \right\} \\
        &\qquad \qquad \qquad - \max_{\*x \in \cX} \left\{ \mu_{t-1}(\*x) + \sqrt{ 2 \log(|\cX| / 2) + \| \*b \|_2^2 } \sigma_{t-1}(\*x) \right\} \mid \cH_{t-1} \biggr] \bigg| \\
        &\leq \left| \EE_{\cD_{t-1}} \left[ \max_{\*x \in \cX} \left\{ \sqrt{ 2 \log(|\cX| / 2) + \| \*a \|_2^2 } \sigma_{t-1}(\*x) - \sqrt{ 2 \log(|\cX| / 2) + \| \*b \|_2^2 } \sigma_{t-1}(\*x) \right\} \mid \cH_{t-1} \right] \right| \\
        &\overset{(a)}{\leq} \left| \sqrt{ 2 \log(|\cX| / 2) + \| \*a \|_2^2 } - \sqrt{ 2 \log(|\cX| / 2) + \| \*b \|_2^2 } \right| \\
        &\overset{(b)}{\leq} \left| \| \*a \|_2 - \| \*b \|_2 \right| \\
        &\overset{(c)}{\leq} \| \*a - \*b \|_2 .
    \end{align*}
    The inequalities (a), (b), and (c) are derived by (a) $\sigma_{t-1}(\*x) \leq 1$, (b) the monotone decreasing property of the gradient of $\sqrt{\cdot}$, and (c) the triangle inequality.
    Therefore, from Proposition~\ref{prop:lipshitz_contraction} and $\bigl(h(\*X) \mid \cH_{t-1}\bigr) \overset{\mathrm{d}}{=} \bigl(\EE_{\cD_{t-1}} \left[ v_t(\*x_t) \mid \zeta_t \right] \mid \cH_{t-1} \bigr)$, 
    \begin{align*}
        \myPr_{\zeta_t} \left( \left| \EE_{\cD_{t-1}, \zeta_t}[ v_t(\*x_t)] - \EE_{\cD_{t-1}} \left[ v_t(\*x_t) \right] \right| > c \mid \cH_{t-1} \right)
        \leq 2 \exp \left\{ - c^2 / 2 \right\},
    \end{align*}
    for any $\cH_{t-1}$.
    Therefore, using Proposition~\ref{prop:tail_prob_to_subgaussian}, $\EE_{ \cD_{t-1}, \zeta_t} \left[ v_t(\*x_t) \mid \cH_{t-1} \right] - \EE_{ \cD_{t-1}} \left[ v_t(\*x_t) \mid \cH_t \right] \mid \cH_{t-1}$ is $18$-subgaussian.

    Finally, since we see that $A_1$ is the sum of random variables that satisfy the condition of Azuma's inequality, $A_1$ is $18T$-subgaussian.
    Therefore, from Proposition~\ref{prop:subgaussian_to_tail_prob}, the following inequality holds with probability at least $1 - \delta_1$,
    \begin{align*}
        A_1 \leq 6\sqrt{T \log(1 / \delta_1)}.
    \end{align*}
    Hence, for some fixed $T \geq 1$, we obtain the high-probability bounds of $A_1$ and $A_2$.
    Then, substituting $6 \delta / (\pi^2 T^2)$ into $\delta_1$ and $\delta_2$ and taking the union bounds for all $T \geq 1$, we conclude the proof.
\end{proof}

%%%%%%%%%%%%%%%%%%%%%%%%%%%%%%%%%%%%%%%%%%%%%%%%%%%%%%%%%%%%%%%%%%%%%%%%%%%%%%%%
\subsection{Proof of Theorem~\ref{theo:HPCER_IRGPUCB_continuous}}
\label{app:proof_conditional_hpcr_continuous}
%%%%%%%%%%%%%%%%%%%%%%%%%%%%%%%%%%%%%%%%%%%%%%%%%%%%%%%%%%%%%%%%%%%%%%%%%%%%%%%%

We show the following theorem:
\begin{reptheorem}{theo:HPCER_IRGPUCB_continuous}
    Let $f \sim \cG \cP (0, k)$, where $k$ is a positive semidefinite kernel and $k(\*x, \*x) \leq 1$, and Assumption~\ref{assump:continuous_X} holds.
    Assume that $\zeta_t$ follows a shifted exponential distribution with $s_t = 2d \log(bdr t^2 \bigl( \sqrt{\log (ad)} + \sqrt{\pi} / 2\bigr)) - 2 \log 2$ and $\lambda = 1/2$ for any $t \geq 1$.
    Then, by running IRGP-UCB with $\zeta_t$, the conditional expected regret can be bounded with probability at least $1 - \delta$ as follows: 
    \begin{align*}
        &\Pr \left( \forall T \geq 1, \EE_{f, \{\epsilon_t\}_{t \geq 1}} \left[ \sum_{t=1}^T f(\*x^*) - f(\*x_t) \mid \{\zeta_t\}_{t \geq 1} \right] \leq U \bigl(T, \delta \bigr) \right) \geq 1 - \delta, 
    \end{align*}
    where
    \begin{align*}
        U(T, \delta) 
        &= \frac{\pi^2}{6} + 6\sqrt{T \log\left(\frac{\pi^2 T^2}{3\delta} \right)} 
        % \\ &\quad 
        + \sqrt{C_1 \gamma_T \left( T s_T + T + 2\sqrt{T \log\left(\frac{\pi^2 T^2}{3\delta} \right)} + 2\log\left(\frac{\pi^2 T^2}{3\delta} \right)\right) },
    \end{align*}
    and $C_1 \coloneqq 2 / \log(1 + \sigma^{-2})$. 
\end{reptheorem}
\begin{proof}
    % Purely for the sake of analysis, we used a set of discretization $\cX_t \subset \cX$ for $t \geq 1$.
    %
    % For any $t \geq 1$, let $\cX_t \subset \cX$ be a finite set with each dimension equally divided into $\tau_t = bdr t^2 \bigl(\sqrt{\log (ad)} + \sqrt{\pi} / 2 \bigr)$.
    %
    We consider a finite discretized input set $\cX_t \subset \cX$ for $t \geq 1$ for theoretical analysis, though $\cX_t$ is not related to the actual algorithm.
    Let $\cX_t = \{\frac{r}{\tau_t + 1}, \dots, \frac{\tau_t r}{\tau_t + 1} \}^d \subset \cX$, where each coordinate takes values from a uniform grid of $\tau_t = bdr u_t \bigl( \sqrt{\log (ad)} + \sqrt{\pi} / 2 \bigr)$ points.
    Thus, $|\cX_t| = \tau_t^d$.
    In addition, we define $[\*x]_t$ as the nearest point in $\cX_t$ of $\*x \in \cX$.

    As with the proof of Theorem~\ref{theo:BCR_IRGPUCB_continuous}, we obtain
    \begin{align*}
        &\EE_{f, \{\epsilon_t\}_{t \geq 1}} \left[ \sum_{t=1}^T f(\*x^*) - f(\*x_t) \mid \{\zeta_t\}_{t \geq 1} \right] \\
        &\leq
        \EE_{f} \left[ \sum_{t=1}^T f(\*x^*) - f([\*x^*]_t) \right]
        + \EE_{f, \{\epsilon_t\}_{t \geq 1}} \left[ \sum_{t=1}^T \max_{\*x \in \cX_t} f(\*x) - f(\*x_t) \mid \{\zeta_t\}_{t \geq 1} \right].
    \end{align*}
    Then, the first term can be bounded from above by $\pi^2 / 6$ as in the proof of Theorem~\ref{theo:BCR_IRGPUCB_continuous}.
    For the second term, we see that
    \begin{align*}
        &\EE_{f, \{\epsilon_t\}_{t \geq 1}} \left[ \sum_{t=1}^T \max_{\*x \in \cX_t} f(\*x) - f(\*x_t) \mid \{\zeta_t\}_{t \geq 1} \right] \\
        &=
        \sum_{t=1}^T \EE_{\cD_{t-1}} \biggl[ \EE_{f \mid \cD_{t-1}} \left[\max_{\*x \in \cX_t} f(\*x)\right] - \EE_{\zeta_t}\left[\max_{\*x \in \cX_t} v_t(\*x)\right] 
        + \EE_{\zeta_t}\left[\max_{\*x \in \cX_t} v_t(\*x)\right] - \EE_{\zeta_t}[v_t(\*x_t)] \\
        & \qquad \qquad \qquad + \EE_{\zeta_t}[v_t(\*x_t)] - v_t(\*x_t) + v_t(\*x_t) - \mu_{t-1}(\*x_t) \mid \{\zeta_t\}_{t \geq 1} \biggr] \\
        &\leq
        \sum_{t=1}^T \EE_{\cD_{t-1}} \biggl[ \EE_{\zeta_t}[v_t(\*x_t)] - v_t(\*x_t) + \zeta_t^{1/2} \sigma_{t-1}(\*x_t) \mid \{\zeta_t\}_{t \geq 1} \biggr],
    \end{align*}
    where we use
    \begin{align*}
        \EE_{f \mid \cD_{t-1}} \left[\max_{\*x \in \cX_t} f(\*x)\right] &\leq \EE_{\zeta_t}\left[\max_{\*x \in \cX_t} v_t(\*x)\right], && (\because \text{Lemma~\ref{lem:bound_RGPUCB}}) \\
        \EE_{\zeta_t}\left[\max_{\*x \in \cX_t} v_t(\*x)\right] &\leq \EE_{\zeta_t}\left[v_t(\*x_t)\right]. && \left(\because \*x_t = \argmax_{\*x \in \cX} v_t(\*x) \right)
    \end{align*}
    Then, the rest of the proof is almost the same as Theorem~\ref{theo:HPCER_IRGPUCB_discrete}.
\end{proof}

\section{Proofs for High-Probability Regret Bounds of IRGP-UCB}
\label{app:proof_hpcr}

% %%%%%%%%%%%%%%%%%%%%%%%%%%%%%%%%%%%%%%%%%%%%%%%%%%%%%%%%%%%%%%%%%%%%%%%%%%%%%%%%
% \subsection{Proof of Theorem \ref{theo:HPCR_IRGPUCB_discrete}}
% \label{app:proof_hpcr_discrete}

Here, we show the proof of the following Theorem~\ref{theo:HPCR_general_GPUCB_discrete}:
\begin{reptheorem}{theo:HPCR_general_GPUCB_discrete}
    Let $f \sim \cG \cP (0, k)$, where $k$ is a positive semidefinite kernel and $k(\*x, \*x) \leq 1$, and $\cX$ be finite.
    Pick $\delta \in (0, 1)$.
    Assume that $\zeta_t = 2 \log ( 1 / U_t )$, where $\{ U_t \}_{t \geq 1}$ is a sequence of mutually independent random variables that satisfy $\EE[U_t] = \frac{6 \delta}{|\cX| t^2 \pi^2}$ and $\Pr(U_t \in (0, 1)) = 1$ for all $t \geq 1$.
    Then, by running the randomized GP-UCB with $\zeta_t$, the cumulative regret can be bounded with probability at least $ 1 - \delta$ as follows: 
    \begin{align*}
        &\Pr \left( \forall T \geq 1, R_T \leq 2\sqrt{C_1 \gamma_T \sum_{t=1}^T \zeta_t } \right) \geq 1 - \delta, 
    \end{align*}
    where $C_1 \coloneqq 2 / \log(1 + \sigma^{-2})$.
\end{reptheorem}
\begin{proof}
    We can transform the regret as follows:
    \begin{align*}
        R_T &= \sum_{t=1}^T f(\*x^*) - f(\*x_t) \\
        &= \sum_{t=1}^T f(\*x^*) - \mu_{t-1}(\*x_t) + \mu_{t-1}(\*x_t) - f(\*x_t) \\
        &\leq 2 \sum_{t=1}^T \zeta_t^{1/2} \sigma_{t-1}(\*x_t) && (\because \text{Lemma~\ref{lem:high_prob_bound_RGPUCB}}) \\
        &\leq 2 \sqrt{ \sum_{t=1}^T \zeta_t \sum_{t=1}^T \sigma_{t-1}^2(\*x_t)} && (\because \text{Cauchy--Schwartz inequality}) \\
        &\leq 2 \sqrt{ C_1 \gamma_T \sum_{t=1}^T \zeta_t }. && (\because \text{Lemma~5.4 in \citet{Srinivas2010-Gaussian}})
    \end{align*}
\end{proof}

Next, we show the following corollary:
\begin{repcorollary}{theo:HPCR_IRGPUCB_discrete}
    Let $f \sim \cG \cP (0, k)$, where $k$ is a positive semidefinite kernel and $k(\*x, \*x) \leq 1$, and $\cX$ be finite.
    Pick $\delta \in (0, 1)$.
    Assume that $\zeta_t$ follows a shifted exponential distribution with $s_t = 2 \log (|\cX| t^2 \pi^2 / (6 \delta))$ and $\lambda = 1/2$ for any $t \geq 1$.
    Then, by running IRGP-UCB with $\zeta_t$, the cumulative regret can be bounded with probability at least $ 1 - \delta$ as follows: 
    \begin{align*}
        &\Pr \left( \forall T \geq 1, R_T \leq 
        2 \sqrt{C_1 \gamma_T \left( T s_T + T + 2\sqrt{T \log\left( \frac{\pi^2 T^2}{3\delta}\right)} + 2\log\left( \frac{\pi^2 T^2}{3 \delta}\right)  \right) } \right) \geq 1 - \delta, 
    \end{align*}
    where $C_1 \coloneqq 2 / \log(1 + \sigma^{-2})$.
\end{repcorollary}
\begin{proof}
    From the assumption that $s_t = 2 \log (|\cX| t^2 \pi^2 / (6\delta))$ and $\lambda = 1/2$, we obtain that $U_t \sim {\rm Uni}(0, \frac{6\delta}{|\cX|t^2 \pi^2})$ and $\EE[U_t] = \frac{3\delta}{|\cX|t^2 \pi^2}$, where $\zeta_t = 2 \log (1 / U_t)$ as in Lemma~\ref{lem:high_prob_bound_RGPUCB}.
    Moreover, from Theorem~\ref{theo:HPCR_general_GPUCB_discrete}, we obtain
    \begin{align*}
        &\Pr \left( \forall T \geq 1, R_T \leq 2\sqrt{C_1 \gamma_T \sum_{t=1}^T \zeta_t } \right) \geq 1 - \frac{\delta}{2}. 
    \end{align*}
    Furthermore, since $\sum_{t=1}^T \{ \zeta_t - s_t \}$ follows a chi-square distribution with $T$ degrees of freedom, using the union bound and Lemma~\ref{lem:Laurent}, we can obtain
    \begin{align*}
        \Pr \left( 
        \forall T \geq 1,
            \sum_{t=1}^T \zeta_t
            \leq 
            T s_T + T + 2\sqrt{T \log \frac{\pi^2 T^2}{3 \delta} } + 2\log \frac{\pi^2 T^2}{3 \delta}
            \right) 
        \geq 1 - \frac{\delta}{2}.
    \end{align*}
    Therefore, taking the union bound concludes the proof.
\end{proof}

\section{Proof for Expected Regret Bound of GP-UCB with Constant Confidence Parameter}
\label{app:proof_GPUCB_BCR_LB}

% %%%%%%%%%%%%%%%%%%%%%%%%%%%%%%%%%%%%%%%%%%%%%%%%%%%%%%%%%%%%%%%%%%%%%%%%%%%%%%%%

First, we derive the following lemma to show Theorem~\ref{theo:GPUCB_BCR_LB}:
\begin{lemma}
    Let $\{\epsilon_i\}_{i \geq 1}$ be a mutually independent sequence of standard normal random variables, that is, $\epsilon_i \sim \cN(0, 1)$ for all $i \geq 1$.
    Then, there exists a positive constant $C$ such that
    \begin{align*}
        \Pr\left( \forall t \leq T, \frac{\sum_{i=1}^t \epsilon_i}{t} \geq -1 \right) \geq C.
    \end{align*}
    \label{lem:sumGauss_lower}
\end{lemma}
\begin{proof}
    From the properties of the probability and the union bound, we can see that
    \begin{align*}
        \Pr\left( \forall t \leq T, \frac{\sum_{i=1}^t \epsilon_i}{t} \geq -1 \right)
        &= 1 - \Pr\left( \exists t \leq T, \frac{\sum_{i=1}^t \epsilon_i}{t} < -1 \right) \\
        &\geq 1 - \sum_{t = 1}^T \Pr\left( \frac{\sum_{i=1}^t \epsilon_i}{t} < -1 \right) \\
        &= 1 - \sum_{t = 1}^T \Pr\left( \frac{\sum_{i=1}^t \epsilon_i}{t} > 1 \right),
    \end{align*}
    where the last equality is obtained from the symmetry of the Gaussian distribution.
    Since $\epsilon_i \sim \cN(0, 1)$ for all $i \geq 1$, the average $\sum_{i=1}^t \epsilon_i / t \sim \cN(0, 1 / t)$.
    Therefore, by using Lemma~\ref{lem:Gauss_tail_bound}, we obtain that
    \begin{align*}
        \Pr\left( \frac{\sum_{i=1}^t \epsilon_i}{t} > 1 \right) \leq \frac{1}{2} \exp \left(- \frac{t}{2} \right).
    \end{align*}
    Since $\bigl(\frac{1}{2} \exp \left(- \frac{t}{2} \right) \bigr)_{t \geq 1}$ is a geometric series with common ratio $\exp (- \frac{1}{2})$ and initial term $\frac{1}{2}\exp (- \frac{1}{2})$, its sum of infinite series converges as follows:
    \begin{align*}
        \sum_{t = 1}^T \Pr\left( \frac{\sum_{i=1}^t \epsilon_i}{t} > 1 \right) 
        &\leq \sum_{t = 1}^T \frac{1}{2} \exp \left(- \frac{t}{2} \right) \\
        &\leq \sum_{t = 1}^\infty \frac{1}{2} \exp \left(- \frac{t}{2} \right) \\
        &= \frac{1}{2 \exp(1/2) \bigl(1 - \exp (- 1/2) \bigr) } \approx 0.771.
    \end{align*}
    Consequently, we can obtain the following lower bound of the probability:
    \begin{align*}
        \Pr\left( \forall t \leq T, \frac{\sum_{i=1}^t \epsilon_i}{t} \geq -1 \right)
        &\geq 1 - \sum_{t = 1}^T \Pr\left( \frac{\sum_{i=1}^t \epsilon_i}{t} > 1 \right) \\
        &\geq 1 - \frac{1}{2 \exp(1/2) \bigl(1 - \exp (- 1/2) \bigr) } \approx 0.229.
    \end{align*}
\end{proof}

Then, we show the following theorem:
\begin{reptheorem}{theo:GPUCB_BCR_LB}
    Assume that $\cX = \{ \*x^{(1)}, \*x^{(2)} \}$, $\epsilon_t \sim \cN(0, 1)$ for all $t \geq 1$, and 
    \begin{align*}
        \left(
            \begin{array}{c}
                 f(\*x^{(1)}) \\
                 f(\*x^{(2)})
            \end{array}
        \right) \sim \cN \left(
            \left(
            \begin{array}{c}
                 0 \\
                 0
            \end{array}
        \right),
        \left(
            \begin{array}{cc}
                 1 & \rho \\
                 \rho & 0.99
            \end{array}
        \right)
        \right),
    \end{align*}
    where $\rho < 1$ is a covariance parameter.
    Then, if GP-UCB with any constant confidence parameter $\beta_t^{(1/2)} = c > 0$ runs, then BCR grows linearly, that is,
    \begin{align*}
        {\rm BCR}_T &= \EE \left[ \sum_{t=1}^T f(\*x^*) - f(\*x_t) \right] = \Omega(T).
    \end{align*}
\end{reptheorem}
\begin{proof}
    We consider the following event:
    \begin{align*}
        E_T = \left\{ f(\*x^{(1)}) \geq \frac{2 \max\{1, c \}}{1 - \rho} + 1, f(\*x^{(2)}) > f(\*x^{(1)}) + 1, \text{ and } \sum_{i=1}^t \epsilon_i / t \geq -1 \text{ for all } t \geq 1 \right\}.
    \end{align*}
    From Lemma~\ref{lem:sumGauss_lower} and the independentness between $f$ and $(\epsilon_i)_{i \geq 1}$, the probability of this event $\Pr(E_T) > 0$.
    Then, the BCR can be bounded from below as follows:
    \begin{align}
        {\rm BCR}_T &= \EE \left[ \sum_{t=1}^T f(\*x^*) - f(\*x_t) \right] \\
        &= \EE \left[ \sum_{t=1}^T f(\*x^*) - f(\*x_t) \mid E_T \right] \Pr(E_T) + \EE \left[ \sum_{t=1}^T f(\*x^*) - f(\*x_t) \mid E_T^{\rm c} \right] \Pr(E_T^{\rm c}) \\
        &\geq \EE \left[ \sum_{t=1}^T f(\*x^*) - f(\*x_t) \mid E_T \right] \Pr(E_T),
    \end{align}
    where $E_T^{\rm c}$ is a complementary event of $E_T$.

    We show that, when $E_T$ holds, $\*x_t = \*x^{(1)}$ for all $t \leq T$.
    First, since the variance of $\*x^{(1)}$ is larger than that of $\*x^{(2)}$ in the prior distribution, the first chosen input $\*x_1 = \*x^{(1)}$.
    Then, we show $\*x_{t+1} = \*x^{(1)}$ for all $t \geq 1$.
    Fix $t \geq 1$ and assume that $\*x_i = \*x^{(1)}$ for all $i \leq t$.
    The posterior mean can be derived as follows:
    \begin{align*}
        \mu_t(\*x^{(1)}) &= \*1_t^\top \left( \*1_t \*1_t^\top + \*I_t \right)^{-1} \*y_t, \\
        \mu_t(\*x^{(2)}) &= \rho \*1_t^\top \left( \*1_t \*1_t^\top + \*I_t \right)^{-1} \*y_t,
    \end{align*}
    where $\*1_t = (1, \dots, 1)^\top \in \RR^t$ and $\*I_t \in \RR^{t \times t}$ is the identity matrix.
    From the Woodbury formula, we obtain
    \begin{align*}
        \left( \*1_t \*1_t^\top + \*I_t \right)^{-1} = \*I_t - \frac{\*1_t \*1_t^\top}{t+1}.
    \end{align*}
    Therefore, the posterior means can be transformed as
    \begin{align*}
        \mu_t(\*x^{(1)}) &= \sum_{i=1}^t y_i - \frac{t}{t+1} \sum_{i=1}^t y_i = \frac{t}{t+1} \frac{\sum_{i=1}^t y_i}{t}, \\
        \mu_t(\*x^{(2)}) &= \frac{\rho t}{t+1} \frac{\sum_{i=1}^t y_i}{t}.
    \end{align*}
    Thus, the difference between the posterior means is bounded from below as
    \begin{align*}
        \mu_t(\*x^{(1)}) - \mu_t(\*x^{(2)}) 
        &= (1 - \rho) \frac{t}{t+1} \frac{\sum_{i=1}^t y_i}{t} \\
        &\geq \frac{1 - \rho}{2} \frac{t f(\*x^{(1)}) + \sum_{i=1}^t \epsilon_i}{t} && (\because t / (t + 1) \geq 1 / 2), \\
        &\geq \frac{1 - \rho}{2} \frac{\frac{2t \max\{1, c \}}{1 - \rho} + 1 - 1}{t} && (\because E_T \text{ holds}) \\
        &= \max\{1, c \} \\
        &\geq c \\
        &= \beta_t^{(1/2)} \\
        &\geq \beta_t^{(1/2)} \sigma_t(\*x^{(2)}) && (\because \sigma_t(\*x^{(2)}) \leq 1) \\
        &\geq \beta_t^{(1/2)} \sigma_t(\*x^{(2)}) - \beta_t^{(1/2)} \sigma_t(\*x^{(1)}).
    \end{align*}
    Hence, we see that $\*x_{t+1} = \*x^{(1)}$ since $\mu_t(\*x^{(1)}) + \beta_t^{(1/2)} \sigma_t(\*x^{(1)}) \geq \mu_t(\*x^{(2)}) + \beta_t^{(1/2)} \sigma_t(\*x^{(2)})$.

    Since $E_T$ holds, the optimal point $\*x^* = \*x^{(2)}$.
    However, as shown above, $\*x_t = \*x^{(1)}$ for all $t \geq 1$ if $E_T$ holds.
    Therefore, $\EE \left[ \sum_{t=1}^T f(\*x^*) - f(\*x_t) \mid E_T \right] \geq T$.
    Consequently, since $\Pr(E_T) > 0$, we can conclude that
    \begin{align*}
        {\rm BCR}_T &= \Omega(T).
    \end{align*}
\end{proof}
\section{Auxiliary Lemmas}
\label{app:lemmas}

%%%%%%%%%%%%%%%%%%%%%%%%%%%%%%%%%%%%%%%%%%%%%%%%%%%%%%%%%%%%%%%%%%%%%%%%%%%%%%%%
For convenience, we here show the assumption again:
\begin{repassumption}{assump:continuous_X}
    Let $\cX \subset [0, r]^d$ be a compact and convex set, where $r > 0$.
    Assume that the kernel $k$ satisfies the following condition on the derivatives of sample path $f$.
    There exist the constants $a, b > 0$ such that,
    \begin{align*}
        \Pr \left( \sup_{\*x \in \cX} \left| \frac{\partial f}{\partial \*x_j} \right| > L \right) \leq a \exp \left( - \left(\frac{L}{b}\right)^2 \right),\text{ for } j \in [d],
    \end{align*}
    where $[d] = \{1, \dots, d\}$.
\end{repassumption}

%%%%%%%%%%%%%%%%%%%%%%%%%%%%%%%%%%%%%%%%%%%%%%%%%%%%%%%%%%%%%%%%%%%%%%%%%%%%%%%%
Then, from Assumption~\ref{assump:continuous_X}, we obtain several lemmas.
First, we show an upper bound of the supremum of partial derivatives.
This result is tighter than Lemma~12 in \citep{Kandasamy2018-Parallelised}.
Note that, in the proof of Lemma~12 in \citep{Kandasamy2018-Parallelised}, there is a typo that $\Pr (L \geq t)$ is bounded above by $a \exp\{ t^2 / b^2 \}$, which should be $a d \exp\{ - t^2 / b^2 \}$.
\begin{lemma}
    Let $f \sim \cG \cP (0, k)$ and Assumption~\ref{assump:continuous_X} holds.
    Let the supremum of the partial derivatives $L_{\rm max} \coloneqq \sup_{j \in [d]} \sup_{\*x \in \cX} \left| \frac{\partial f}{\partial x_j} \right|$.
    Then, $\EE[L_{\rm max}]$ can be bounded above as follows:
    \begin{align}
        \EE[L_{\rm max}] \leq b \bigl( \sqrt{\log (ad)} + \sqrt{\pi} / 2 \bigr).
    \end{align}
    \label{lem:L_max_bound}
\end{lemma}
\begin{proof}
    From the assumption, using the union bound, we obtain a bound on the probability,
    \begin{align}
        \Pr\left( L_{\rm max} > L \right) \leq \sum_{j=1}^d a \exp \left( - \left(\frac{L}{b}\right)^2 \right) = a d \exp \left( - \left(\frac{L}{b}\right)^2 \right). \label{eq:L_max_prob}
    \end{align}
    Then, using expectation integral identity, we can bound $\EE[L_{\rm max}]$ as follows:
    \begin{align}
        \EE[L_{\rm max}]
        &= \int_0^{\infty} \Pr\left( L_{\rm max} > L \right) {\rm d} L \\
        &\leq \int_0^{\infty} \min \{ 1, ad e^{- (L / b)^2} \} {\rm d} L && \bigl(\because \text{Eq.~\eqref{eq:L_max_prob}} \bigr) \\
        &= b \sqrt{\log (ad)} + \int_{b\sqrt{\log (ad)}}^{\infty} ad e^{- (L / b)^2} {\rm d} L \\
        &= b \sqrt{\log (ad)} + a b d \sqrt{\pi} \int_{b\sqrt{\log (ad)}}^{\infty} \frac{1}{\sqrt{2\pi (b^2/2)}} e^{- (L / b)^2} {\rm d} L \\
        &= b \sqrt{\log (ad)} + a b d \sqrt{\pi} \left( 1 - \Phi \left( \frac{b\sqrt{\log (ad)}}{ b / \sqrt{2}} \right) \right) \\
        &\leq b \sqrt{\log (ad)} + \frac{ b \sqrt{\pi}}{2}, && \bigl(\because \text{Lemma~\ref{lem:Gauss_tail_bound}} \bigr)
    \end{align}
    where $\Phi$ is a cumulative distribution function of the standard normal distribution.
\end{proof}

%%%%%%%%%%%%%%%%%%%%%%%%%%%%%%%%%%%%%%%%%%%%%%%%%%%%%%%%%%%%%%%%%%%%%%%%%%%%%%%%
Next, we show the bound for the difference of discretized outputs:
\begin{lemma}
    Let $f \sim \cG \cP (0, k)$ and Assumption~\ref{assump:continuous_X} holds.
    Let $\cX_t = \{\frac{r}{\tau_t + 1}, \dots, \frac{\tau_t r}{\tau_t + 1} \}^d \subset \cX$, where each coordinate takes values from a uniform grid of $\tau_t = bdr u_t \bigl( \sqrt{\log (ad)} + \sqrt{\pi} / 2 \bigr)$ points.
    %
    % be a finite set with each dimension equally divided into $\tau_t = bdr u_t \bigl( \sqrt{\log (ad)} + \sqrt{\pi} / 2 \bigr)$ for any $t \geq 1$.
    %
    Then, we can bound the expectation of differences,
    \begin{align}
        \sum_{t=1}^T \EE \left[ \sup_{\*x \in \cX} f(\*x) - f( [\*x]_t ) \right] \leq \sum_{t=1}^T \frac{1}{u_t},
    \end{align}
    where $[\*x]_t$ is the nearest point in $\cX_t$ of $\*x \in \cX$.
    \label{lem:discretized_error}
\end{lemma}
\begin{proof}
    From the construction of $\cX_t$, we can obtain the upper bound of L1 distance between $\*x$ and $[\*x]_t$ as follows:
    \begin{align}
        \sup_{\*x \in \cX}\| \*x - [\*x]_t \|_1
        &\leq \frac{dr }{bdr u_t \bigl( \sqrt{\log (ad)} + \sqrt{\pi} / 2\bigr) } \\
        &= \frac{1}{b u_t \bigl( \sqrt{\log (ad)} + \sqrt{\pi} / 2\bigr)}.
        \label{eq:sup_L1dist_discretization}
    \end{align}
    Note that this discretization does not depend on any randomness and is fixed beforehand.

    Then, we obtain the following:
    \begin{align}
        \sum_{t=1}^T \EE \left[ \sup_{\*x \in \cX} f(\*x) - f( [\*x]_t ) \right] 
        &\leq \sum_{t=1}^T \EE \left[ L_{\rm max} \sup_{\*x \in \cX} \| \*x - [\*x]_t \|_1 \right] \\
        &\leq \sum_{t=1}^T \EE \left[ L_{\rm max} \right] \frac{1}{b u_t \bigl( \sqrt{\log (ad)} + \sqrt{\pi} / 2\bigr)} && \bigl(\because \text{Eq.~\eqref{eq:sup_L1dist_discretization}} \bigr) \\
        &\leq \sum_{t=1}^T \frac{1}{u_t}. && \bigl(\because \text{Lemma~\ref{lem:L_max_bound}} \bigr)
    \end{align}
\end{proof}

%%%%%%%%%%%%%%%%%%%%%%%%%%%%%%%%%%%%%%%%%%%%%%%%%%%%%%%%%%%%%%%%%%%%%%%%%%%%%%%%

We used the following useful lemma:
\begin{lemma}[in Lemma 5.2 of \citep{Srinivas2010-Gaussian}]
    For $c > 0$, the survival function of the standard normal distribution can be bounded above as follows:
    \begin{align}
        1 - \Phi(c) \leq \frac{1}{2} \exp (- c^2 / 2).
    \end{align}
    \label{lem:Gauss_tail_bound}
\end{lemma}
% \begin{proof}
%     Let $R \sim \cN(0, 1)$.
%     %
%     Then, we obtain the following:
%     \begin{align}
%         \Pr(R > c) &= \int_{c}^\infty \frac{1}{\sqrt{2 \pi}} \exp ( - r^2 / 2) {\rm d}r \\
%         &= \exp (- c^2 / 2) \int_{c}^\infty \frac{1}{\sqrt{2 \pi}} \exp ( - r^2 / 2 + c^2 / 2) {\rm d}r \\
%         &= \exp (- c^2 / 2) \int_{c}^\infty \frac{1}{\sqrt{2 \pi}} \exp ( - (r - c)^2 / 2 - rc + c^2) {\rm d}r \\
%         &= \exp (- c^2 / 2) \int_{c}^\infty \frac{1}{\sqrt{2 \pi}} \exp ( - (r - c)^2 / 2 - c(r - c)) {\rm d}r \\
%         &\leq \exp (- c^2 / 2) \int_{c}^\infty \frac{1}{\sqrt{2 \pi}} \exp ( - (r - c)^2 / 2) {\rm d}r && \left( \because \text{$e^{-c(r-c)} \leq 1$ since $r \geq c > 0$} \right) \\
%         &= \frac{1}{2} \exp (- c^2 / 2).
%     \end{align}
%     %
%     % 参考までに, 以下のバウンドも成り立つ ($c \geq \sqrt{\pi / 2}$ではこちらのほうがタイト):
%     % \begin{align}
%     %     \Pr(R > c) &= \int_{c}^\infty \frac{1}{\sqrt{2 \pi}} \exp ( - r^2 / 2) {\rm d}r \\
%     %     &\leq \frac{1}{\sqrt{2 \pi}} \int_{c}^\infty \frac{r}{c} \exp ( - r^2 / 2) {\rm d}r && \text{($r/c \geq 1$ in $r \geq c$)}\\
%     %     &= \frac{1}{\sqrt{2 \pi}} \frac{1}{c} \exp (- c^2 / 2).
%     % \end{align}
% \end{proof}

%%%%%%%%%%%%%%%%%%%%%%%%%%%%%%%%%%%%%%%%%%%%%%%%%%%%%%%%%%%%%%%%%%%%%%%%%%%%%%%%

\begin{lemma}[Eq~4.3 of \citep{Laurent2000-adaptive}]
    Pick $\delta \in (0, 1)$.
    Suppose that $X$ follows a chi-square distribution with $D$ degrees of freedom.
    Then, the following upper bound for the tail probability holds:
    \begin{align*}
        \Pr\left( X \geq D + 2 \sqrt{Dc} + 2c \right) \leq \exp (- c),
    \end{align*}
    for any $c > 0$.
    Therefore, the following inequality holds with probability at least $1 - \delta$:
    \begin{align*}
        X \leq D + 2 \sqrt{D \log (1 / \delta)} + 2 \log (1 / \delta).
    \end{align*}
    \label{lem:Laurent}
\end{lemma}

%%%%%%%%%%%%%%%%%%%%%%%%%%%%%%%%%%%%%%%%%%%%%%%%%%%%%%%%%%%%%%%%%%%%%%%%%%%%%%%%%%%%%%%

% \tableofcontents
\end{document}